\documentclass[manuscript]{jair}

\setcopyright{cc}
\acmDOI{10.1613/jair.1.22258}

\JAIRAE{Valerio Basile}
\JAIRTrack{Surveys} %
\acmVolume{86}
\acmArticle{48}
\acmMonth{08}
\acmYear{2026}

\usepackage{annotation}
\usepackage{arydshln}
\usepackage{float}
\usepackage{pifont}
\usepackage{wrapfig}
\usepackage{multirow}
\usepackage{url}
\usepackage{graphicx}
\usepackage{hyperref}
\usepackage{cleveref}
\usepackage{booktabs}
\usepackage{natbib} %
\usepackage{apalike}

\AtBeginDocument{%
  \providecommand\BibTeX{{%
    \normalfont B\kern-0.5em{\scshape i\kern-0.25em b}\kern-0.8em\TeX}}}

\begin{document}

\title[Text Generation]{Text Generation: A Systematic Literature Review of Tasks, Evaluation, and Challenges}

\author{Jonas Becker}
\authornote{Corresponding Author.}
\orcid{0009-0006-6438-1211}
\email{jonas.becker@uni-goettingen.de}
\affiliation{%
  \institution{University of Göttingen}
  \streetaddress{Papendiek 14}
  \postcode{37073}
  \country{Germany}
}
\affiliation{%
  \institution{LKA NRW}
  \streetaddress{Völklinger Str. 49}
  \city{Düsseldorf}
  \postcode{40221}
  \country{Germany}
}

\author{Jan Philip Wahle}
\orcid{0000-0002-2116-9767}
\email{wahle@uni-goettingen.de}
\affiliation{%
  \institution{University of Göttingen}
  \streetaddress{Papendiek 14}
  \postcode{37073}
  \country{Germany}
}

\author{Bela Gipp}
\orcid{0000-0001-6522-3019}
\email{gipp@uni-goettingen.de}
\affiliation{%
  \institution{University of Göttingen}
  \streetaddress{Papendiek 14}
  \postcode{37073}
  \country{Germany}
}

\author{Terry Ruas}
\orcid{0000-0002-9440-780X}
\email{ruas@uni-goettingen.de}
\affiliation{%
  \institution{University of Göttingen}
  \streetaddress{Papendiek 14}
  \postcode{37073}
  \country{Germany}
}

\renewcommand{\shortauthors}{Becker, Wahle, Gipp \& Ruas}

\begin{abstract}
Text generation has become more accessible than ever, and the growing interest in these systems, especially those using large language models, has spurred a surge in related publications.
We provide a systematic literature review comprising 257 papers, covering the period from January 2017 to December 2025.
This review categorizes text generation contributions into five main tasks: open-ended text generation, summarization, translation, paraphrasing, and question answering.
For each task in our taxonomy, we review relevant characteristics and key subtasks.
We assess current approaches for evaluating text generation systems, covering model-free, model-based, and human evaluation.
Our investigation shows several task-specific challenges (e.g., missing datasets for multi-document summarization, lack of coherence in story generation, and difficulties in complex reasoning for question answering).
We further discuss nine challenges common to all tasks and sub-tasks in recent text generation papers: bias, reasoning, hallucinations, misuse, privacy, interpretability, transparency, datasets, and computing.
This systematic literature review targets two main audiences: early-career researchers in natural language processing seeking an overview of the field and promising research directions, and senior researchers who need a recent overview of the main tasks, evaluation, challenges, and mitigation strategies.
\end{abstract}

\received{18 March 2026}
\received[accepted]{15 July 2026}

\maketitle
\AddAnnotationRef{}

\section{Introduction} \label{sec:intro}

Human language is a defining feature of human intelligence, enabling communication, reasoning, and shared understanding across individuals and time \citep{Dennett13, Premack04}.
Text generation is a core task in natural language processing that involves producing coherent sequences of natural language tokens.
Similar to how industrialization has automated numerous physical tasks and bodily routines, artificial intelligence (AI) systems are becoming human companions to solve cognitive problems and mental routine tasks \citep{OuyangWJA22, AchiamAAA24, TouvronLIM23, GeminiTeamABW23}.

Major breakthroughs in AI have been achieved when models became capable of modeling natural language, specifically with large language models (LLMs), generating texts that are on par with human writing \citep{UchenduMLZ21, DouFKS22a, WahleRFM22b}.
Together with deep learning architectures, large-scale data, and accessible computing infrastructure, this has unveiled a new paradigm for training AI assistants at scale.
Anyone with internet access can now have their own AI assistant to automate laborious, time-consuming tasks such as drafting emails, filling out forms, or developing software.
This scenario is the result of sustained, collaborative, and incremental efforts by researchers and practitioners worldwide in AI, engineering, statistics, linguistics, and natural language processing (NLP), inter alia, over the years.

In the early days of distributional semantics \citep{Harris54}, and rule-based question answering systems \citep{McKeown82}, models would use structured data, grammatical rules, and templates to construct text \citep{winograd1972automatic,McKeown82}.
Today, most approaches use neural networks and backpropagation via gradient descent to estimate the probability of the next word in a sequence of words \citep{BengioDV00}.
Modern LLMs like GPT-5 \citep{singh2025openai}, LLaMA \citep{TouvronLIM23}, and Gemini \citep{GeminiTeamABW23} solve problems in a text-to-text manner through natural language. 
This problem-solving method is inspired by how humans solve problems, formulating answers that convey a particular meaning, a process known as language production.
In terms of machines, text generation is the process of creating natural-language text.

Text generation, also referred to as natural language generation (NLG), is present in a myriad of tasks, such as generating stories \citep{BrownMRS20b}, engaging in dialogue \citep{LiZ23}, or solving reasoning problems \citep{GeminiTeamABW23}.
Despite this rapid expansion, the literature on text generation remains fragmented across tasks, evaluation practices, and assumptions about what constitutes progress.
As a result, contributions are often difficult to compare, open problems are inconsistently defined, and limitations are revisited in isolation rather than addressed systematically.

A systematic literature review is therefore essential not merely to summarize existing work, but to organize and contextualize contributions, identify recurring challenges and opportunities. By consolidating models, datasets, and research directions.
Such a review can help establish shared understanding and guide future research toward more robust, interpretable, and responsible text generation.

In this paper, we provide an overview of recent text generation research up to December 2025, focusing on publications since the introduction of the Transformer architecture in 2017 \citep{VaswaniSPU17a}, which has since become a central component of NLP.
Our systematic literature review has three main focuses: tasks and sub-tasks (\Cref{sec:textgen}), evaluation metrics (\Cref{sec:eval}), and challenges (\Cref{sec:challenges}).
For this literature review, we exclude investigations of data-to-text and multimodal approaches (e.g., syntactic-semantic representations of images \citep{TianZXW24} and tables \citep{DingX23}). 
This is done given the recent prominence of work on text-to-text tasks and the fact that evaluation and challenges inherently differ between text-to-text and data-to-text systems, exceeding the scope of this review. 
For multimodality, we refer to the surveys by \citet{10.1613/jair.1.12918, 10041115}.
We devise the following key questions to organize our review:
\begin{enumerate}
    \item What constitutes the task of text generation? What are the main sub-tasks?
    \item How are text generation systems evaluated? What are the concomitant limitations?
    \item What are the open challenges in text generation?
    \item What are prominent research directions in text generation?
\end{enumerate}

\begin{figure}[t]
    \centering
    \includegraphics[width=\linewidth]{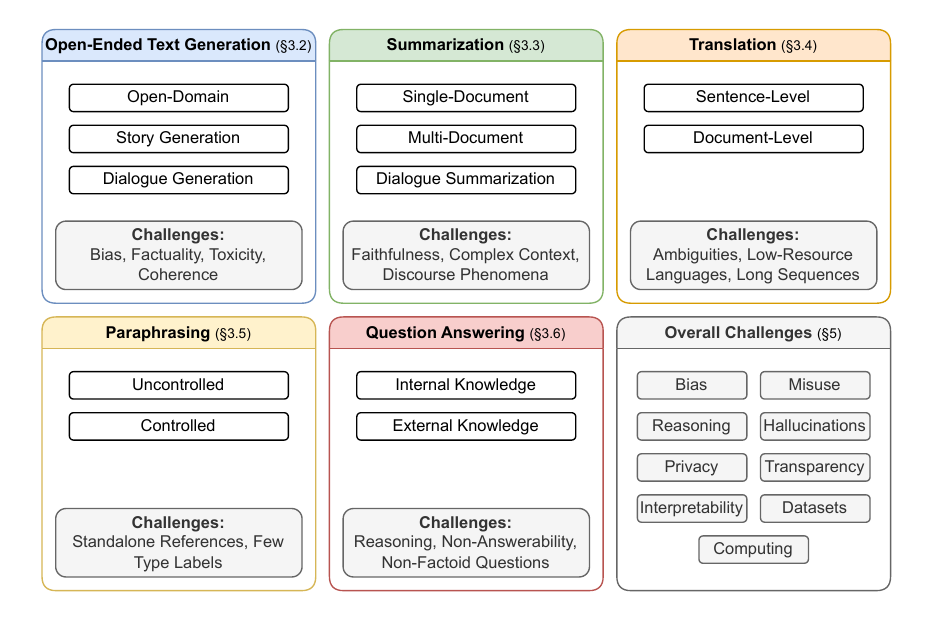}
    \caption{Taxonomy of the main tasks in text generation. For each main task (e.g., translation), we identify common subtasks (e.g., sentence-level, document-level). We discuss the specific challenges of each task and the nine overall challenges in text generation.}
    \label{fig:tasks}
\end{figure}

\Cref{fig:tasks} gives an overview of the most prominent tasks and associated challenges in text generation.
We delimit text generation as the synthetic production of contextually relevant content in written natural language from an underlying non-linguistic representation.
We identify five main tasks through our investigation: \textit{open-ended text generation}, \textit{summarization}, \textit{translation}, \textit{paraphrasing}, and \textit{question answering} (\Cref{sec:textgen}).
For each task, we review its relevant characteristics, sub-tasks, and specific challenges.
Next, we assess commonly employed evaluation methodologies in the field (i.e., model-free and model-based metrics, LLM-as-a-judge, and human evaluation) and discuss their limitations (\Cref{sec:eval}).
In addition, we also identify nine prominent challenges common to all tasks and sub-tasks in recent text generation publications: \textit{bias}, \textit{reasoning}, \textit{hallucinations}, \textit{misuse}, \textit{privacy}, \textit{interpretability}, \textit{transparency}, \textit{datasets}, and \textit{computing} (\Cref{sec:challenges}).
Lastly, we revisit, summarize, and answer our research questions (\Cref{sec:epiloge}).
For reproducibility, we publicly share the details of our methodology (e.g., key phrases, subjective decisions, exclusion criteria, code) and metadata are publicly available\footnote{\url{https://github.com/jonas-becker/text-generation/}\label{github_project}}.

\subsection{Related Literature Reviews}
Since 2017, the field of text generation has seen a marked increase in publications \citep{RamosMR23}.
As a result, several works survey and organize the available literature.
Most of these efforts focus on particular aspects of text generation, such as efficient Transformer architectures \citep{TayDBM20, LinWLQ21}, adaptations of pre-trained models for unconditional text generation \citep{LiTZN24}, the use of reinforcement learning for sequence-to-sequence models \citep{KeneshlooSRR19}, the use of external knowledge to improve the quality of generated text \citep{YuZLH22}, and specific challenges (e.g., hallucination \citep{JiLFY23}, bias \citep{SunGTH19, GallegosRBT24}, human evaluation \citep{ClarkASH21}, security \citep{YaoDXC24}).

Only a few literature reviews provide a broad analysis of text generation.
\citet{10.5555/3241691.3241693} survey the state-of-the-art in text generation in 2018, covering tasks, applications, and evaluation.
\citet{ZhaoZLT23a} investigate the pre-training, fine-tuning, utilization, and evaluation of recent LLMs with more than ten billion parameters in selected tasks.
Similarly, \citet{10433480} review the field of LLMs, covering a variety of NLP tasks including text generation.
\citet{FatimaIKD22} survey 90 publications (2015-2021) on text generation using deep neural networks and organize their approaches, quality metrics, datasets, languages, and applications.
\citet{GoyalKS23} survey the technical application of text generation and concomitant challenges starting from the year 2011.
The survey addresses table-to-text generation and multimodal tasks like image captioning.

Our literature review contributes to the existing literature by its recency, breadth of coverage across tasks, and systematic retrieval of papers.
First, we include a large set of 257 resources resulting from subsequent exploration of the text generation field, covering the literature until December 2025.
Second, unlike other literature reviews, such as \citet{JiLFY23}, we systematically select our primary publications and document each step in the selection process.
Rather than choosing specific subfields preemptively \citep{LiTZN22a, ZhaoZLT23a}, we adopt a bottom-up approach and infer the subtasks, evaluation procedures, and associated challenges in text generation through a systematic search, adding supplementary works at a later stage during writing.
While other literature reviews focus on more technical characteristics (e.g., architectures \citep{GoyalKS23, TayDBM20, LinWLQ21, 10433480}), we provide a comprehensive investigation of text generation sub-tasks, their concomitant solutions, and remaining problems.
Specifically, we differ from \citet{GoyalKS23}, who outline technical procedures for generating text and explain statistical methods, deep learning, transformer-based models, and encoding techniques.
While they also mention some task-specific challenges, they do not provide a comprehensive overview focused on text-to-text tasks, including overall challenges of text generation as we do.
We further add to their contribution by quantifying the impact and application of commonly employed evaluation metrics within the scientific community.
\citet{TayDBM20, LinWLQ21} exclusively review Transformers, whereas we focus specifically on text-generation sub-tasks and use a bottom-up approach to infer commonly employed techniques.
Finally, we dedicate a specific analysis to open challenges in text generation that can be explored in the near future, thereby informing researchers in the field about which research gaps are most promising to pursue.

\section{Methodology} \label{sec:method}

\begin{figure}[t]
    \centering
    \includegraphics[width=\linewidth]{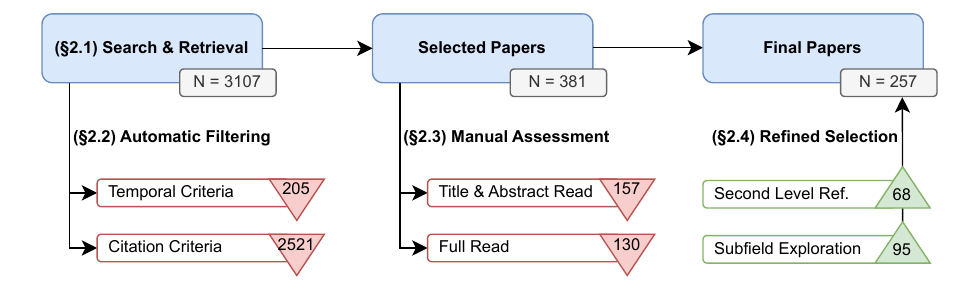}
    \caption{Pipeline of this systematic review. Red-colored triangles show papers we excluded; Green-colored triangles show papers we added. We report the judgments on manually filtered papers in our GitHub repository.}
    \label{fig:filtering_pipeline}
\end{figure}

This systematic literature review is inspired by the guidelines of \citet{KitchenhamC07,Okoli15} and comprises three parts: \textit{search and retrieval}, \textit{automatic filtering}, and \textit{manual assessment}. 
\Cref{fig:filtering_pipeline} provides an overview of the process. 
We provide the source code for reproducing the literature retrieval on GitHub\textsuperscript{\ref{github_project}}.

\subsection{Search and Retrieval} \label{sec:searchandretrieval}

\noindent\textbf{Query construction.} Our search queries use \textit{primary} and \textit{secondary} terms to select candidate papers.
While primary terms aim to match papers in text generation, secondary terms focus our query on specific characteristics (e.g., training).
As the two primary terms, we use \textit{text generation} and \textit{machine-generated text}.
We generate secondary terms by manually inspecting existing field surveys and using suggestions from ChatGPT-4\footnote{version May 12, 2023}. 
We manually review the list of search terms, merge overlapping terms, and conduct consensus voting among the authors in cases of ambiguity to finalize the selection.
We consider the following 14 secondary terms: \textit{paraphrase}, \textit{nlp}, \textit{natural language generation}, \textit{language model}, \textit{evaluation metrics}, \textit{bias}, \textit{privacy}, \textit{controllable}, \textit{creative}, \textit{machine}, \textit{automated}, \textit{task}, \textit{training}, \textit{cost}, \textit{co2 emission}, and \textit{detection}.

\noindent\textbf{Retrieval.} We search Semantic Scholar for papers (including preprints from arXiv) by building all permutations of primary and secondary terms as queries, leaving us with 30 queries (e.g., ``text generation evaluation metrics'').
We restrict our search to Semantic Scholar's field of study to computer science to exclude articles on text generation in other disciplines (e.g., text generation tasks by humans in psychology studies)\footnote{According to Semantic Scholar, the Fields of Study attribute is assigned using a machine-learning classifier based on paper titles and abstracts; the S2FOS classifier achieved 86\% correctness in an end-system human evaluation: \href{https://medium.com/ai2-blog/announcing-s2fos-an-open-source-academic-field-of-study-classifier-9d2f641949e5}{https://medium.com/ai2-blog/announcing-s2fos-an-open-source-academic-field-of-study-classifier-9d2f641949e5}}.
We retrieve the top 100 papers per query and year, ranked by Semantic Scholar's native ranking algorithm\footnote{\url{https://api.semanticscholar.org/api-docs/}}.
Semantic Scholar ranks search results using a learned relevance function that scores how well each document matches the query based on multiple signals, including semantic content, citation information, and metadata.
The system combines an initial text-matching stage (e.g., keyword/semantic similarity) with a learned open-source reranker (s2search\footnote{\url{https://github.com/allenai/s2search}}) that reorders results to prioritize papers most likely relevant to the query.
Our search results in 3107 publications published between 1966---2025.

\subsection{Automatic Filtering} \label{sec:automaticfiltering}

\noindent\textbf{Temporal criteria.}
The Transformer architecture has fundamentally changed the landscape of text generation and now forms the basis of modern language models \citep{VaswaniSPU17a}.
While existing surveys predominantly examine approaches developed before the Transformer era \citep{10.5555/3241691.3241693} or focus on its early adoption \citep{FatimaIKD22}, this review considers work published since the Transformer's introduction in 2017.
By this, we expect to retrieve works that directly or indirectly rely on the Transformer architecture.
This cutoff is defined because most LLMs are Transformer-based, enabling a systematic analysis of the architecture's abilities, limitations, and emerging research opportunities.

\noindent\textbf{Citation criteria.}
Next, we filter the results using Semantic Scholar's \textit{influential citations} \citep{ValenzuelaEscarcegaHE15}.
Influential citations, therefore, provide a proxy for measuring a publication's impact on other publications in the field.
An influential citation traces key works that another work relies on, e.g., measured by citations in specific paper locations, research overlap, or similar work \citep{ValenzuelaEscarcegaHE15}.
To even out the distribution of papers by year, we sample the top five papers per query and year by influential citations.
We select papers with more than two influential citations until 2021 to estimate papers of strong influence in text generation.
We treat recent works differently as older papers tend to have more citations than recent ones, although both can be impactful \citep{AbdallaWRN23, WahleRAG23}.
Thus, we relax our restrictions and do not remove papers with two or fewer influential citations for the years 2022 to 2025. 
After the citation criteria, we are left with 381 works.

\subsection{Manual Assessment} \label{sec:manual_assessment}

\noindent\textbf{Title \& Abstract Read.}
We manually assess the titles and abstracts of all 381 works from 2017 to 2025 to judge their relevance regarding text generation.
We eliminate works that employ text generation within another field of study (e.g., medicine).
Multimodality for text generation requires data-specific solutions (e.g., images \citep{TianZXW24} and tables \citep{DingX23}).
The focus of this review is text generation and tasks that can be solved through natural language.
Therefore, we remove papers about multimodality in the process.
We release a table of our relevance annotations to our GitHub repository\textsuperscript{\ref{github_project}}.

\begin{table}[ht]
    \centering
    \small
        \caption{Number of papers per year after search and retrieval (\Cref{sec:searchandretrieval}), after automatic filtering (\Cref{sec:automaticfiltering}), and after refined selection (\Cref{sec:refinedselection}).}
    \label{tab:yearly_stats_flipped}
    \renewcommand{\arraystretch}{1.2}
    \resizebox{\linewidth}{!}{%
        \begin{tabular}{lrrrrrrrrrr|r}
            \toprule
            Number of papers after step... & \textbf{Prior} & \textbf{2017} & \textbf{2018} & \textbf{2019} & \textbf{2020} & \textbf{2021} & \textbf{2022} & \textbf{2023} & \textbf{2024} & \textbf{2025} & \textbf{Total} \\
            \midrule
            \textbf{§2.1 Search \& Retrieval} & 205 & 65 & 113 & 188 & 334 & 438 & 268 & 52 & 1147 & 297 & \textbf{3107}\\
            \textbf{§2.2 Automatic Filtering} & 0 & 47 & 64 & 48 & 48 & 41 & 41 & 48 & 29 & 15 & \textbf{381}\\
            \textbf{§2.4 Refined Selection} & 34 & 12 & 20 & 26 & 37 & 27 & 28 & 42 & 18 & 13 & \textbf{257}\\
            \bottomrule
        \end{tabular}%
    }
\end{table} %

\noindent\textbf{Full Read.}
We read and summarize each relevant paper, considering its main contributions, applications, tasks, datasets, potential challenges, and limitations.
The publications are tagged with the different main aspects of this review (e.g., dataset, metric) to facilitate their organization. 
We release a sampled table of our notes and tagged terms to our GitHub repository\textsuperscript{\ref{github_project}}.

\subsection{Refined Selection} \label{sec:refinedselection}

\noindent\textbf{Second Level References.}
To complement our review with other relevant work, we add auxiliary papers that are cited by the papers that undergo a full read.
These auxiliary papers are direct references (second level).
For example, this includes cases in which we find that method papers repeatedly use a dataset, architecture, or metric, indicating that the resource is relevant to the field and should be included in this literature review.
We include 68 papers in the process.

\noindent\textbf{Subfield Exploration.}
As we identify key areas of text generation, such as challenges, evaluation methods, and subtasks, we conduct a targeted search within these areas.
This step serves two purposes.
First, it complements citation-based retrieval by capturing recent advances in text generation that have not yet accumulated sufficient citations to be reliably surfaced by automated search engines such as Semantic Scholar.
Second, it ensures comprehensive coverage across text-generation subtasks, evaluation, and challenges. After identifying core thematic areas, a targeted search facilitates the retrieval of relevant studies, such as architectural or task-specific analyses, that may not be captured by broad keyword-based queries alone.
We note that the literature review adopts a bottom-up strategy, in which relevant tasks, evaluation methods, and challenges are systematically identified from earlier stages of the review process.
This targeted subfield exploration is conducted within these inferred categories.
For example, we search for the term "multi-document summarization" on scholarly platforms, such as Google Scholar, because this subtask was previously identified in the retrieved works.
Ultimately, this step improves the recency, per-area coverage, and depth of our literature review.
Adding 95 works in the process, our final selection of reviewed papers comprises 257 works as of December 2025.

\section{Text Generation Tasks} \label{sec:textgen}
Text generation, also called natural language generation, is a computational linguistic task that produces human-like text from a latent (underlying) non-linguistic representation of information \citep{Mcdonald80, Katz80, ReiterD97}. 

In this paper, we treat text generation as a central capability and use it to organize the discussion of downstream tasks. 
Depending on the input and the intended behavior, text generation includes, for instance, generating text from structured data, rewriting existing text \citep{RadevHM02}, producing responses in interactive settings \citep{LiZ23}, and generating domain-specific text such as code \citep{roziere2022leveragingautomatedunittests}. 

Text generation is relevant beyond its technical role. 
It can lower barriers to accessing information and interacting with technology, including for users with higher learning curves \citep{naumanen2007guiding}, and it can support domain work by turning specialized texts into more accessible forms \citep{DagdelenDLW24}. 
At the same time, the same capability can be misused to generate convincing misinformation at scale, for example via social media bots \citep{ZellersHRB20}. 
Whether text generation is considered an opportunity, a risk, or both, understanding the recent developments in this field is of the utmost importance.
Therefore, we define the problem of text generation, its most prominent sub-tasks, concomitant characteristics, and open challenges.

\subsection{Problem}
\citet{Katz80} was one of the first to explore text generation as a process of grouping different feature representations for generating kernel sentences, deciding on transformations based on syntax and themes, and then combining these kernels with appropriate modifications and pronoun adjustments to produce cohesive text.
Later, \citet{ReiterD97} expanded the notion of text generation to signals other than text, such as weather forecast data from graphical weather maps \citep{GoldbergDK94} or the creation of technical documentation via a knowledge base \citep{REITERML95}.
Modern definitions advocate for a process that leverages knowledge from computational linguistics and artificial intelligence to automatically generate natural, cohesive, and plausible texts that meet defined requirements \citep{LiTZN22a}. %
In our literature review, we delimit text generation as the synthetic production of contextually relevant content in written natural language using an underlying non-linguistic representation.
While the input to a text generation system during training might be natural language, it is not directly converted to a textual output.
The output is derived from an underlying non-linguistic representation (e.g., embeddings) displaying meaning. 

At the time of writing, text generation is often performed via LMs \citep{LiTZN22a, BrownMRS20b}.
LMs provide a probabilistic description of natural language and maximize the probability of the generated sequence, producing the most probable text relevant to the context \citep{BengioDV00, JurafskyM24}.
Language modeling and text generation are distinct yet interconnected fields.
While text generation focuses on producing content, language modeling leverages the probabilistic structure of language.
However, text generation does not necessarily require LMs, rule-based approaches from \citet{Katz80} and non-autoregressive methods \citep{XiaoWGL23} can also generate text. %

Our literature review identifies five prominent areas related to text generation, namely \textit{open-ended text generation}, \textit{summarization}, \textit{translation}, \textit{paraphrasing}, and \textit{question answering} (\Cref{fig:tasks}).
Open-ended text generation is a task where newly generated text is iteratively conditioned on the previous
context so that the final output appears coherent and fluent \citep{LiHFL23, venkatraman-etal-2025-collabstory}.
This also included modern, instruction-tuned AI assistants such as ChatGPT \citep{AchiamAAA24, OuyangWJA22}.
Summarization is the generation of a text from one or more texts to convey information in a shorter format \citep{RadevHM02}.
Translation means converting a source text in language A to a target language B \citep{LewisLGG20a}. %
Paraphrasing is the task of generating text that has (approximately) identical meaning but uses different words or structures \citep{DolanB05a, VilaBMR15a, WahleGR23a}.
Question answering takes a question as input text and outputs a streamlined answer or a list of possible answers based on background knowledge \citep{CaiWZY20}.
We investigate these areas based on their specific subtasks and challenges.
We mention relevant datasets selectively but point to our GitHub repository for an elaborate list of available corpora\textsuperscript{\ref{github_project}}.

\subsection{Open-Ended Text Generation}
\noindent\textbf{Task definition.} Open-Ended Text Generation (OETG) generates new text constrained by an existing source text (e.g., prompt) so that the final output appears coherent and fluent \citep{LiHFL23, HeZWK23a}.
Applications of OETG range from open-domain response generation by AI assistants \citep{AchiamAAA24} to complex story generation \citep{LiangTLZ23}.

\noindent\textbf{Main characteristics.} Compared to other text generation tasks like summarization, OETG provides more freedom during content generation due to the highly variable output length.
For this, LMs that maximize the probability distribution over word sequences are used to generate fluent text.
Conversely, human language mainly encompasses semantic and pragmatic characteristics beyond grammatical and syntactic rules \citep{GehrmannSR19a}.
Nevertheless, the fluency of machine-generated text can mislead human judgments of authorship \citep{AdelaniMFN20}.
We note that some of the following tasks and sub-tasks may also be considered OETG in an LLM-prompted setup.
Previous advances in instruction-following \citep{OuyangWJA22} have led to greater task unification under general-purpose systems like ChatGPT \citep{AchiamAAA24}.
We will not reiterate the possibility of prompting an OETG system like ChatGPT \citep{AchiamAAA24} for each task.
This is crucial for this review, as we focus on the task-specific modeling, datasets, and challenges.

\noindent\textbf{Challenges.} Since the modern days of LMs \citep{BrownMRS20b, DevlinCLT19a} and their capacity to produce fluent texts, other desired features emerged for OETG.
As sequences generated by OETG systems can grow large, the risk of producing logical inconsistencies increases.
This makes coherence a key challenge to solve in OETG \citep{GuanMFL21, LiangTLZ23}.
In addition, recent LMs are subject to bias, prejudice, and toxicity from human language. 
Thus, controlling attributes like sentiment \citep{Kawamae23, AdelaniMFN20, LiuSLS21} during OETG is an active research topic \citep{YangLLY23}.

\subsubsection{Sub-tasks and Datasets in Open-Ended Text Generation} \hfill

\noindent We identify three popular sub-tasks of OETG, namely \textit{open-domain} OETG, \textit{story generation}, and \textit{dialogue generation}. 
Niche sub-tasks, such as review generation \citep{CostaODL18, NiM18} and data-to-document generation \citep{LuDXZ23}, often use open-domain OETG methods because they generalize well to these problems.

\noindent\textbf{Open-domain:} %

\noindent\textit{Description.}
Given minimal or underspecified input, the model generates coherent, contextually appropriate text across a wide range of subjects and scenarios, allowing for high variability in content, style, and structure.
The task is not constrained to a predefined topic, format, or domain.

\noindent\textit{Modeling.}
Open-domain OETG differs from traditional NLP tasks in that it is not defined by a fixed domain or output structure, but rather by the system’s ability to generate coherent and appropriate text under specified conditions \citep{li-etal-2023-contrastive}.
Thus, open-domain OETG systems must generalize not only across topics and domains but also across tasks, discourse structures, and styles, which are provided implicitly by natural-language prompts rather than by explicit task specifications.
To achieve this, the dominant approaches for open-domain OETG use LLMs like GPT \citep{AchiamAAA24, BrownMRS20b, RadfordWCL} or LLaMA \citep{TouvronLIM23}, which support a wide range of downstream objectives through natural language prompting.
This includes applications such as ChatGPT \citep{AchiamAAA24, OuyangWJA22}, where users prompt the system to solve a variety of tasks (e.g., instruction following \citep{OuyangWJA22}, question generation \citep{MullaG23}).
Prompting reframes diverse tasks as instances of conditional text generation, in which instructions serve as soft constraints rather than fixed optimization targets, thereby avoiding the need for task-specific architectures or supervision.

Improvements in open-domain OETG are largely attributed to training strategies, including large-scale pre-training on diverse data, instruction tuning, and alignment to human preferences, rather than to the introduction of task-specific model architectures \citep{OuyangWJA22}.
Large-scale pre-training on datasets such as Common Crawl\footnote{https://commoncrawl.org/} \citep{BrownMRS20b, TouvronLIM23} or knowledge bases \citep{YuZLH22} improves coverage across domains and formats, but does not guarantee correctness, particularly in rare or complex cases.
Thus, evaluation plays a central role in open-domain OETG.
To evaluate pre-trained LLMs for OETG, \citet{he-etal-2021-tgea} provide an error-annotated dataset and benchmark tasks for text generation.
Their benchmark covers errors from a taxonomy of 24 error types, organized into main and subtypes in linguistics and knowledge.
Frequent errors of generated text are commonsense errors, inappropriate combinations, missing words, discourse errors, and redundancies.
Open-domain OETG inherently trades increased coverage for greater exposure to rare and complex cases, where errors are more likely to arise \citep{liu-etal-2022-challenges}.
Consequently, effective open-domain OETG systems must balance broad generalization with mechanisms that promote linguistic and factual correctness.

\noindent\textit{Challenges of open-domain OETG.} 
Open-domain OETG requires balancing multiple competing objectives—such as coherence, relevance, factuality, and diversity—that are often in tension with one another \citep{HoltzmanWBZS20}.
Improvements along one dimension (e.g., diversity) may degrade others (e.g., coherence or factual accuracy), complicating both modeling and evaluation.
Contrastive decoding selects each next token by favoring what a strong model prefers over what a weaker ``generic'' model prefers, steering generation away from bland high-frequency continuations \citep{li-etal-2023-contrastive}.
This usually yields more specific, less repetitive text.
Technology companies often propose the most capable open-domain models as they have the resources to train them on large amounts of data \citep{AbdallaWRN23}.
Many of these models are closed-source.
This makes the creation and reproducibility of modern LLMs particularly challenging, as many companies tend not to disclose technical details about their models, such as used data or hyperparameters \citep{AchiamAAA24}.
Few companies follow an open-source approach when it comes to LLM development \citep{TouvronLIM23}.
This can hinder the scientific study of OETG, as these systems are difficult to reproduce and might not generalize to widely available open-source models.
We note that open-domain OETG is also challenged by biased datasets, interpretability issues, limited evaluation, and other factors.
As these challenges are not exclusive to OETG and are prevalent across several text generation tasks, we address them generally in \Cref{sec:eval} and \Cref{sec:challenges}.

\noindent\textbf{Story Generation:}

\noindent\textit{Description.}
Story generation is a conditional text generation task that requires producing narrative text that is coherent, logically consistent, and engaging, typically involving characters, events, and a meaningful temporal progression for the reader \citep{GuanMFL21, LiangTLZ23}.

\noindent\textit{Modeling.}
We find that most approaches employ an LM with additional modules to enhance planning, temporal consistency, or coherence.
Unlike short-form text generation, story generation requires maintaining a consistent narrative structure, character development, and causal progression over extended contexts.
Thus, several works explicitly introduce planning mechanisms.
Planning the generation process hierarchically can enhance perceived creativity \citep{Peng22}.
To achieve planning, \citet{YangTPK22} employ a module that produces a story premise with a setting, characters, and outline. 
Additionally, they use a dedicated module to make local edits, ensuring that the generated story remains consistent.
Other approaches focus on improving coherence through prompt construction and conditioning.
\citet{FanLD18} use a convolutional LM to generate a specially designed prompt and enhance the consistency of story generation with a text-to-text model.
Their work also introduces the WritingPrompts dataset, which contains over 300k human-written prompt–story pairs.
CollabStory uses several LLMs that collaborate at the segment level to form a coherent storyline \citep{venkatraman-etal-2025-collabstory}.

\noindent\textit{Challenges of story generation.} The production of long and coherent stories with a consistent narrative and developing characters (i.e., unique and changing properties over the temporal shift in the story) is one of the main challenges in story generation \citep{AlabdulkarimLP21}.
Commonly employed decoding algorithms like beam search produce incoherent and repetitive text for story generation \citep{WiherMC22}.
While effective for constrained tasks (e.g., translation), beam search performs poorly in open-ended story generation where many plausible continuations exist.
This is because beam search maximizes sequence likelihood, favoring high-probability and generic continuations.
In open-ended story generation, this can lead to degeneration phenomena such as repetition loops and loss of global coherence.
To mitigate this, \citet{LuWWJ22} propose a decoding algorithm with lookahead heuristics that enables foresight planning.
This requires that the constraints of the conditioned story can be formulated as logical expressions.

\noindent\textbf{Dialogue generation:}

\noindent\textit{Description.}
Dialogue generation is a multi-turn conditional text generation task of modeling verbal or written communication among two or more participants, producing contextually appropriate and coherent utterances in a conversational exchange \citep{ZhangWLG20a, LiZ23}.
While a two-party setup (i.e., training data and architecture) is often used in AI assistants, a multi-party setup is more common in, e.g., movie dialogue.

\noindent\textit{Modeling.}
Dialogue contains one-to-many relationships, where a single dialogue context may correspond to multiple appropriate replies.
As bidirectional LMs (e.g., BERT \citep{DevlinCLT19a}) and autoregressive LMs (e.g., GPT \citep{AchiamAAA24, BrownMRS20b, RadfordWCL}) are not directly optimized on dialogue datasets, and the one-to-many relationships make fine-tuning inherently difficult, dialogue generation is modeled differently than open-domain OETG or story generation.
\citet{BaoHWW20} and \citet{ZhangSGC20a} avoid fine-tuning by further pre-training the LM on conversational data.
They jointly integrate uni- and bidirectional processing and model the one-to-many relationships with a latent variable.
For pre-training and testing conversational models, mostly synthetic data exists.
For example, the AMI corpus \citep{MccowanCKA05} comprises 100 hours of transcribed meeting dialogue, partly real and mostly staged between actors.
The proposal of more authentic, human-authored dialogue at a sufficiently large scale could be a meaningful contribution to the field of dialogue generation.

\noindent\textit{Challenges of dialogue generation.} We identify key challenges on the technical level and the availability of datasets.
Both two- and multi-party dialogue generation are challenging tasks due to semantics, consistency, and interactivity when applied to open-domain problems \citep{HuangZG20}.
\citet{HuangZG20} identify that these challenges are influenced by long context, character personalities, sentiment, and dialogue policies.
\citet{10691641} study persona-driven dialogues inspired by role-playing games, using a card-based framework that includes scene settings and character personas.
When guided by detailed scene settings, their small model (seven billion parameters) could craft context-aware dialogues for even unseen characters and scenarios.
While research on two-party dialogue is common, multi-party dialogue generation remains underexplored \citep{KannEKD22}, possibly because natural multi-party dialogue often occurs in confidential settings, such as business meetings.
This makes the available datasets for training or evaluating multi-party dialogue generation scarce.
Investigating the dynamics of multi-party dialogue could support the understanding of the dynamics behind turn-taking \citep{Skantze21} and multi-agent interaction \citep{WangMWG23, becker-etal-2025-mallm}.

\subsection{Summarization}
\noindent\textbf{Task definition.} Summarization describes the generation of a short text, distilling one or more references \citep{RadevHM02}. 
These references are usually more than double the length of the produced summaries \citep{RadevHM02}.
Many works differentiate between \textit{extractive}, \textit{abstractive}, and \textit{hybrid} summarization \citep{FabbriKMX21a, BrazinskasLT20, TaunkSPS23}. 
Extractive summarization concatenates spans of the input text to produce the final output text \citep{MoratanchC17}.
Abstractive summarization generates new phrases and sentences that can differ from the original text \citep{ZhangZSL20a}.
Moreover, hybrid approaches aim to combine the benefits of both extractive and abstractive methods \citep{TaunkSPS23}.

\noindent\textbf{Main characteristics.} The characteristics of the task are different when looking at either extractive or abstractive summarization.
Extractive systems use statistics to identify relevant text and are usually simple to implement, e.g., by fine-tuning an LM to select relevant text snippets \citep{LiuWNC20} or ranking sentences from a source \citep{XiaoC19a}. 
Abstractive systems, conversely, use semantic features like word embeddings directly \citep{ZhangZSL20a} to determine the meaning of texts and summarize them accordingly.
Hence, abstractive summarization can generate text with low lexical overlap with the source, whereas extractive summarization always has overlap in the extracted text spans \citep{PaulusXS17, ZhangZSL20a}.

\noindent\textbf{Challenges.} Extractive summarization often produces duplicated information and inconsistent summaries sensitive to the statistical features of the source text \citep{MunotS14} while abstractive methods suffer from large contexts when summarizing large documents \citep{GambhirG17}.
In addition, abstractive systems lack faithfulness, often hallucinating \citep{MaynezNBM20}.

\subsubsection{Sub-tasks and Datasets in Summarization} \hfill

\noindent According to our investigations, the most popular summarization sub-tasks are \textit{single-document summarization}, \textit{multi-document summarization}, and \textit{dialogue summarization}. 

\noindent\textbf{Single-Document Summarization (SDS):}

\noindent\textit{Description:} SDS is a sequence-to-sequence text generation task that aims to produce an accurate and concise summary from a single input document, e.g., a news article \citep{ZhangCXW19, ZhangZSL20a}.

\noindent\textit{Modeling.}
Extractive SDS can be performed using encoders such as BERT, a bidirectional masked LM \citep{LiuL19a}.
Copy mechanisms employed by a pointer-generator network leverage relevant tokens from the source as part of the output \citep{KumarRL19}. 
A decoder equipped with a pointer-generator network makes a soft choice at each decoding step between copying from the source and generating from the vocabulary \citep{ZhangCXW19}.
This way, the encoder can predict out-of-vocabulary words by selecting appropriate tokens from the source text.
\citet{XiaoC19a} show that separately accounting for local and global context can improve summary quality.
Abstractive summarization has seen a surge of techniques due to its ability to generate lexically independent summaries \citep{WangYTZ18, ZhangCXW19}. 
Some efforts combine extractive and abstractive methods for summarization to address the long computation times associated with purely abstractive approaches for long documents \citep{TaunkSPS23}.
Today, LLMs are used for abstractive summarization based on a query or topic \citep{yang2023exploringlimitschatgptquery}.
Prompted LLMs can match the performance of traditional fine-tuning approaches, but the possibility of factual errors or biased summaries exists.
Meanwhile, LLMs can adapt their output to different target audiences (e.g., expertise, writing style), but still lack the stylistic diversity of human-authored text \citep{pu-demberg-2023-chatgpt}.

Most SDS datasets we identify originate from the news domain \citep{NarayanCL18a, NallapatiZdG16a}.
This may be due to the fact that these datasets can conveniently leverage headlines or first sentences from news websites as summaries \citep{NarayanCL18a} and are thus comparatively easy to create.
For example, the XSum dataset \citep{NarayanCL18a} includes 227k single-sentence summaries of BBC articles.
The CNN/Daily Mail dataset \citep{NallapatiZdG16a} is a corpus with 312k multi-sentence summaries obtained by concatenating human summary bullets.
The reliance on the news domain yields data that is genre-specific and not necessarily representative of an appropriate summary for other domains.

\noindent\textit{Challenges of SDS.} Very long documents are inherently more challenging to summarize than shorter ones, as the relevance of more text sections needs to be taken into account.
The quadratic time and memory complexity of commonly employed Transformers hinders their direct use for the task \citep{VaswaniSPU17a}. 
This limitation is mitigated by specialized architectures like Longformer \citep{BeltagyPC20b}, which employ a linearly scaling attention mechanism.
Besides this, abstractive summarization systems often lack faithfulness to the source text and tend to hallucinate information \citep{MaynezNBM20}.
Some work addresses the faithfulness of pretrained LMs for document-grounded text generation (e.g., summarization), which we cover in \Cref{sec:challenges_factuality}.

\noindent\textbf{Multi-Document Summarization (MDS):}

\noindent\textit{Description.} MDS is a sequence-to-sequence text generation task in which a model synthesizes information from multiple topic-related documents, often originating from different sources, to produce a single coherent and non-redundant executive summary \citep{MaZGW23}.

\noindent\textit{Modeling.} Common approaches for MDS specify distinct mechanisms to select or attend to relevant documents in a large collection.
\citet{MaynezAG23a} identify several of these strategies for architecture design, ranging from ensemble networks or hierarchical approaches to graph neural networks or fine-tuning of pre-trained LMs.
\citet{LiuXTL19} represent cross-document relationships via an attention mechanism.
If faced with many (hundreds or more) documents, an LM can be employed with retrieval augmented generation (RAG) to combine document vectorization with retrieval of relevant text blocks \citep{NEURIPS2020_6b493230}.
We find only few datasets for MDS.
The Multi-News dataset \citep{FabbriLSL19a} provides 46k news articles with an average number of 3.5 documents per summarized cluster.
DiverseSumm includes 245 news stories, with each story comprising 10 news articles \citep{huang-etal-2024-embrace}.

\noindent\textit{Challenges of MDS.} MDS suffers from similar challenges as SDS but comes with an additional set of issues.
Cross-document relations and conflicting, duplicate, and complementary information increase the challenges of mapping similar entities, events, and other related information \citep{MaZGW23, GambhirG17}.
Modern LLMs often hallucinate information in multi-document settings \citep{belem-etal-2025-single}.
They are also limited in their coverage of information across documents \citep{huang-etal-2024-embrace}.
We also find that the diversity of available MDS datasets is low, possibly due to the costs of creating them.

\noindent\textbf{Dialogue Summarization:}

\noindent\textit{Description.} Dialogue summarization is a sequence-to-sequence text generation task that condenses a multi-party, multi-turn dialogue into a concise and informative summary while preserving key decisions, intents, and salient points \citep{10.1613/jair.1.16674}.
This sub-task is also important for meeting summarization goals \citep{FengFQQ21}, as it can reduce costs and time spent on manual note-taking.

\noindent\textit{Modeling.} Dialogue contains one-to-many relationships, where a single dialogue context may correspond to multiple appropriate replies.
\citet{ZhuXZH20} capture the unique dialogue structure by a hierarchical approach.
They employ a word-level Transformer that processes the sequence of a single turn, while a turn-level Transformer processes the information of all turns in a meeting.
\citet{FengFQQ21} leverage a relational graph encoder to model discourse relations.
Mixture of Experts is with role-oriented routing via an LLM-based module to summarize a dialogue, with each expert processing different information from the dialogue \citep{tian-etal-2024-dialogue}. 
This avoids anticipation bias and the potential loss of information inherent to a standard single-LLM approach.
Most datasets for dialogue summarization consist of synthetic data.
The AMI corpus \citep{MccowanCKA05} comprises 279 hours of synthetic meeting dialogue textualized from speech with annotations of extractive and abstractive summaries.
Its participants play different roles in a team responsible for hardware design.
The SAMSum corpus \citep{gliwa-etal-2019-samsum} includes 16k chat dialogues paired with human-written abstractive summaries.
The dialogues are deliberately created by linguists, and the summaries are written in the third person.

\noindent\textit{Challenges of dialogue summarization.} Dialogue summarization is difficult because of discourse phenomena like coreference errors (e.g., ambiguous pronouns) or discourse link errors (e.g., temporal ordering) \citep{PagnoniBT21}.%
This is why researchers use distinct architectures to account for these discourse phenomena, such as relational graph encoders \citep{FengFQQ21} or hierarchical processing \citep{ZhuXZH20}.
We identify that datasets in dialogue and meeting summarization are scarce, especially when looking for non-synthetic data.
We suppose that this is because of confidentiality issues between the participants and the discussed topics.
Training and evaluation on synthetic data could cause unexpected problems, such as a performance drop in real conversations, because naturally occurring dialogue might differ structurally or linguistically.

\subsection{Translation}
\noindent\textbf{Task definition.} Translation, also referred to as machine translation, is a task in which a (textual) source in language A is converted into a target language B \citep{LewisLGG20a}. %

\noindent\textbf{Main characteristics.} Unlike many natural language processing tasks, translation requires modeling two languages simultaneously and learning systematic correspondences between them.
The publications identified by our search concern text-to-text problems (e.g., German-to-English translation).

\noindent\textbf{Challenges.} One of the main challenges in translation is to guarantee that the generated translation conveys the same meaning as its source (i.e., avoiding semantic shift).
Most contributions in translation often focus on high-resource languages (e.g., English and Chinese).
Consequently, training and evaluation are more difficult for low-resource languages due to the lack of accessible data covering them \citep{TaunkSPS23}.
The mismatch between train and test data (i.e., the model being trained on data different from the data it is evaluated or used on) can significantly impact the performance of any translation system \citep{CurreyMD20, ArtetxeGBF23}.
To mitigate this problem, \citet{ArtetxeGBF23} adapt the human training data by back-translation (e.g., Portoguese to English to Portoguese), making it more similar to a machine-generated model input during test time.
They assume that applying machine translation twice induces a similar error and thus reduces the gap between train and test data.
A well-selected dataset can also help reduce the train/test mismatch \citep{VilarFCL23}.

\subsubsection{Sub-tasks and Datasets in Translation} \hfill

\noindent Among our retrieved papers, the most popular translation sub-tasks are \textit{sentence-level} and \textit{document-level} translation.
Most datasets in translation consider short sequences for their tasks (\textit{sentence-level}) as they are easier to annotate than longer sequences (\textit{document-level}).
Thus, current systems are often more capable of handling short sequences as they are largely trained on corresponding data.
However, \textit{document-level} information might help to solve word ambiguity as the longer context can allow better handling of discourse phenomena like coreference, omissions, and coherence \citep{ZhangZ20a}.

\noindent\textbf{Sentence-Level:}

\noindent\textit{Description.} Sentence-level translation is a sequence-to-sequence text generation task in which maps an individual source sentence to its equivalent in the target language, maintaining semantic meaning and grammatical correctness.

\noindent\textit{Modeling.}
Sentence-level translation involves sentence-aligned datasets, typically sampled from full-sized documents. 
Systems must process relatively short sequences.
As the complexity of short sequences is low, corresponding translation systems are abundant and capable.
Works on sentence-level translation propose a variant of the encoder-decoder Transformer, predicting tokens with a classifier over a large database of translation examples \citep{KhandelwalFJZ21}, or prompt existing pre-trained LLMs \citep{ZhangHB23} to perform machine translation.
The most popular datasets in the translation field come from the WMT General Machine Translation task \citep{KocmiABB23}. 
Since 2008, there have been regular releases and updates to the WMT dataset, which comprises several other corpora \citep{CallisonBurchFKM08, KocmiABB23}.
Each corpus covers a different set of languages (e.g., Chinese, German, Japanese) and domains (e.g., e-commerce, conversation, news), with a few thousand examples per language.
Together, these models and benchmarks have established sentence-level translation as a largely saturated setting, providing a foundation for exploring more challenging scenarios, such as document-level translation.

\noindent\textit{Challenges of sentence-level translation.} Sentence-level translation is specifically limited due to a lack of context.
For example, when referring to out-of-context personalities by pronouns or when encountering words that can have multiple translations depending on the language \citep{ZhangZ20a}.
Word ambiguities are difficult to handle when only a single sentence is provided as a source. 
In addition, translation models trained on short sequences perform significantly worse when tested on long sequences (60+ words) \citep{KoehnK17, 10.1162/tacl_a_00730}.
Thus, success in semantic preservation for longer sequences is especially influenced by the variability, context, and ambiguity of different languages.
Further challenges for modern LLMs in translation include inference efficiency, translation of low-resource languages during pretraining, and evaluation according to human-aligned translation quality \citep{10.1162/tacl_a_00730}.

\noindent\textbf{Document-Level:}

\noindent\textit{Description.} Document-level translation is a sequence-to-sequence text generation task that leverages in-document context to resolve ambiguities, maintain coherence, and ensure consistency across sentences that are difficult to handle in isolated sentence-level translation \citep{MarufSH22}.

\noindent\textit{Modeling.}
The exact length of the segmented sequences heavily differs between works, ranging from small paragraphs to complete books.
Thus, we reference efforts to translate text into sequences longer than sentence-level as document-level translation.
Most document-level translation approaches leverage the long-context modeling capabilities of recent LLMs \citep{WangLJZ23}.
In general, LLMs are more suitable for translating documents than single sentences \citep{10.1162/tacl_a_00730}.
Data for document-level translation is scarce.
Many of the available datasets are paid resources that require a fee to access, such as the NIST corpora \citep{NISTMultimodalInformationGroup10, NISTMultimodalInformationGroup13}.
They are relatively small (100 samples per language) but provide reference translations for longer news articles. 
NIST datasets also only cover a few languages and are not uniformly structured.
For example, NIST 2005 \citep{NISTMultimodalInformationGroup10} contains 4-way reference translations of Newswire articles in Chinese and Arabic (100 samples per language), while NIST 2012 \citep{NISTMultimodalInformationGroup13} contains 222 documents with Chinese-to-English pairs from Newswire and web data.
Our investigation suggests that the lack of human datasets for this sub-task entails high upfront costs for research, as it requires creating high-quality translations of longer documents.

\noindent\textit{Challenges of document-level translation.}
Research on document-level translation remains limited, largely due to the combined challenges posed by long input sequences and the scarcity of large-scale, annotated datasets.
Discourse phenomena such as coreference, omissions, and coherence can occur in a manner that is specific to long sequences, leading to performance differences between translating shorter and longer sequences \citep{ZhangZ20a}.
While sentence-level models struggle to capture such phenomena, modern LLMs can handle more context at the cost of interpretability \citep{WangLJZ23}.
Further studies of document-level translation could yield interpretable improvements in handling ambiguities and discourse phenomena.

\subsection{Paraphrasing}
\noindent\textbf{Task definition.} Paraphrases are texts expressing (approximately) identical meanings that use different words or structures \citep{DolanB05a, VilaBMR15a, WahleGR23a}.
Thus, paraphrasing aims towards the generated text having a high semantic similarity to the source text.

\noindent\textbf{Main characteristics.} In paraphrasing, the output length is not a relevant criterion, which is different from other tasks like translation.
Thus, a paraphrased text can be the same length, shorter, or longer than the original text.
Paraphrases can be generated in several ways, such as by replacing or adding words, making grammatical changes, or changing the order of words \citep{KovatchevMS18a}. 

\noindent\textbf{Challenges.} A key challenge of paraphrasing is the lack of suitable evaluation metrics.
The similarity of paraphrases is difficult to measure, with different lexical or semantic changes taking place.
Most authors use BLEU or other n-gram-based metrics to test their systems' performance \citep{Grusky23a}.
As \citet{ShenLJS22a} point out, this has issues because their correlation with human judgments is low, and there are many ways to paraphrase a text.
Consequently, multiple references for each data point would facilitate the training of paraphrasing systems \citep{ZhengMH18}, but we find that most corpora still only include one reference \citep{KovatchevMS18a, WangHF17a}.
Model-based metrics offer an alternative to overlap-based evaluation, but come at the cost of reduced interpretability and potential model bias \citep{HeZWK23a}.
Machine-generated paraphrases are semantically closer to their original source text when compared to human paraphrases \citep{BeckerWRG23}.
This indicates a gap between human and machine paraphrases that is not captured by model-free quality metrics.
Additionally, some paraphrasing datasets are thematically balanced, whereas others exhibit concentrated or unbalanced coverage of domains such as politics \citep{BeckerWRG23}.

\subsubsection{Sub-tasks and Datasets in Paraphrasing} \hfill

\noindent We identify two major sub-fields of paraphrasing research, namely \textit{uncontrolled} paraphrasing and \textit{controlled paraphrasing}.
As \textit{uncontrolled} paraphrasing does not come with additional constraints on the generation process, it is inherently characterized by its output variability.
The other branch of paraphrasing studies aims to \textit{control} the variability of paraphrases during generation.
As recent models demonstrate satisfactory performance on \textit{uncontrolled} paraphrasing tasks, the focus has shifted toward developing systems that provide greater \textit{control} to the user.
Thus, \textit{uncontrolled} paraphrasing is well-established compared to \textit{controlled} paraphrasing.

\noindent\textbf{Uncontrolled:} 

\noindent\textit{Description.} Uncontrolled paraphrasing is a sequence-to-sequence text generation task in which a model rewrites an input text into a semantically equivalent alternative, without enforcing specific lexical, syntactic, or stylistic constraints on the output.

\noindent\textit{Modeling.}
Common approaches use an encoder-decoder architecture that first creates a semantic representation of the input text and then outputs a paraphrase \citep{EgonmwanC19a}.
\citet{GargPMS21} fine-tune an autoregressive LM to generate paraphrases.
Uniquely, their approach does not rely on parallel paraphrase datasets.
It instead uses lexical control signals to guide how words or phrases should change in the paraphrase.
A reward with reinforcement learning encourages semantic similarity to the original sentence while promoting lexical and syntactic diversity.
Several datasets are available for paraphrasing considering uncontrolled scenarios.
For example, TURL \citep{LanQHX17b} comprises 2.8m multi-rater annotated sentence pairs from Twitter news.
Quora Question Pairs \citep{WangHF17a} comprises 400k similar questions from the Quora forum with binary paraphrase annotations.

\noindent\textit{Challenges of uncontrolled paraphrasing.} Since many possible paraphrases can be produced, the n-gram-based comparison to a single reference is questionable.
To encourage diversity in paraphrases, we, therefore, suggest the use of multiple references per source text \citep{LanQHX17b, ZhengMH18}.
However, these are often not provided in today's paraphrasing datasets.
We suggest that new corpora should come with multiple reference paraphrases as a standard.

\noindent\textbf{Controlled:}

\noindent\textit{Description.} Controlled paraphrasing is a conditional sequence-to-sequence text generation task in which a model produces semantically equivalent text while adhering to predefined control signals, such as paraphrase type (e.g., modal verb changes) \citep{WahleGR23a, VilaMR14a, KovatchevMS18a}, target characteristics like quality \citep{BandelASS22a}, or structural constraints such as syntactic form \citep{ChenTWG19a, KumarAVT20a}.

\noindent\textit{Modeling.}
To generate paraphrases in a controlled manner, most prior work focuses on the encoder side of text-to-text systems.
\citet{WahleGR23a} show that paraphrases of a specific type can be generated using autoregressive models like BART \citep{LewisLGG20a} or PEGASUS \citep{ZhangZSL20a} by conditioning the encoder on task-specific prompts or labels.
Several approaches introduce explicit structural control.
\citet{ChenTWG19a} and \citet{KumarAVT20a} employ semantic and syntactic encoders for syntax-controllable paraphrase generation, while \citet{YangBXZ22} encode constituency trees with Transformers to capture parent–child and sibling relations, constraining outputs to a given syntactic template.
Beyond structural control, \citet{BandelASS22a} directly regulate paraphrase quality by conditioning generation on semantic similarity as well as syntactic and lexical distance.
This enables the model to trade off meaning preservation and lexical and syntactic variation during decoding.
Finally, \citet{ZhangDLY21a} leverage manifold sentiment information to guide paraphrase generation.

The ETPC dataset \citep{KovatchevMS18a} comes with the extended paraphrase typology.
This includes granular annotations of the categories morphology-based, lexicon-based, lexico-syntactic-based, syntax-based, discourse-based, and other changes.
Though it only includes 5k samples and shows a balanced distribution of covered topics \citep{BeckerWRG23}, other datasets usually lack such granular type annotations.
However, some other datasets focus on one specific paraphrase type like sentence splitting \citep{BothaFAB18} or entailments \citep{WilliamsNB18a}. 
Overall, we observe that datasets are abundant for specific paraphrase types, such as sentence splitting, whereas other types, such as polarity changes, syntax control, or sentiment control, require more specialized datasets. 
An additional avenue is to adapt non-paraphrasing research on controllability \citep{GhoshCLM17, HuangZJS20, YangLLY23, SeoJJH23, YangK21}.

\noindent\textit{Challenges of controlled paraphrasing.} While uncontrolled paraphrasing is a well-established task, we find that controlled paraphrasing tasks such as paraphrase type generation are incipient \citep{WahleGR23a}.
Thus, many architectural characteristics that make controlled paraphrasing systems superior remain unclear, as they also depend on the controlled attribute.
We could not identify many datasets that provide a comprehensive selection of annotated paraphrase types or controlled characteristics like sentiment.
Therefore, we encourage other researchers in the field to construct comprehensive datasets that yield controllable attributes, such as paraphrase types \citep{KovatchevMS18a}.
This would foster further research on controlled paraphrasing.

\subsection{Question Answering}
\noindent\textbf{Task definition.} Question answering (QA) takes a question as input and outputs a streamlined answer or a list of possible answers based on background knowledge \citep{CaiWZY20}.
This background knowledge can either be provided as a supplementary document, a knowledge base, or leveraged from training data when no additional context is provided.

\noindent\textbf{Main characteristics.} Background knowledge of a QA system can be \textit{internal} or \textit{external} \citep{YuZLH22}.
\textit{Internal knowledge} takes place within the training data and input text(s), including but not limited to keywords, topics, linguistic features, and internal graph structure.
\textit{External knowledge} acquisition occurs when knowledge is provided from outside sources, such as a knowledge base, knowledge graph, or grounded text.
While there exist QA systems \citep{CaiWZY20} for topic-restricted domains (e.g., math QA), most research focuses on open-domain QA, i.e., QA systems with capabilities that are not constrained on a topic because they generalize better to practical problems. %

\noindent\textbf{Challenges.} The difficulties of QA stem from open-domain evaluation and the interpretability of the system's answers.
Evaluation of generated answers is difficult for very abstractive settings like open-domain QA where the source does not directly indicate what the correct answer could be \citep{DurmusHD20}.
This contrasts with multiple-choice settings, in which evaluation is comparatively simple, e.g., via regular expression text matching.
To bridge the gap of interpretability, researchers draw inspiration from how humans synthesize a chain of thought for reasoning.
Although Chain-of-Thought Prompting \citep{WeiWSB24} and Solo Performance Prompting \citep{WangMWG23} are competent approaches to improve reasoning in QA systems, their implementation is not trivial because they require a specific prompt design and large models that are capable of mimicking human-like reasoning this way.

\subsubsection{Sub-tasks and Datasets in Question and Answering} \hfill

\noindent According to our investigations, the most popular QA sub-tasks are \textit{internal knowledge-grounded QA} (e.g., open-domain QA) and \textit{external knowledge-grounded QA} (e.g., reading comprehension).

\noindent\textbf{Internal Knowledge:}

\noindent\textit{Description.} Internal-knowledge question answering is a text generation task in which a model produces an answer solely based on the input question and knowledge implicitly stored in its parameters or other underlying representations \citep{YuZLH22}.
The system can not leverage additional resources to answer a question.
Some works refer to internal knowledge as parametrized knowledge \citep{ma-etal-2025-one}.

\noindent\textit{Modeling.}
Corresponding systems are mostly employed with pre-trained LLMs that utilize the internal knowledge memorized from large web corpora \citep{TanMLL23}.
Such internalized knowledge supports open-domain question answering by enabling the model to retrieve and combine relevant information without explicit access to external databases at inference time.
Large and diverse pre-training corpora like Common Crawl\footnote{\url{https://commoncrawl.org/}} increase the amount of internal knowledge that can be leveraged for QA, which is particularly important for handling rare entities, long-tail queries, and heterogeneous question formulations.
Recent work suggests that the factual knowledge encoded in a model's internal computations can significantly exceed what the model actually generates in its outputs, revealing that some correct answers may be present internally yet not always produced by repeated sampling methods \citep{gekhman2025insideouthiddenfactualknowledge}. Moreover, this hidden knowledge gap constrains the practical gains achievable by increasing test-time compute in open-domain QA, because some of the model's factual information remains inaccessible when relying solely on token-level generation.
Beyond pre-training, domain-specific datasets can be used for fine-tuning to adapt the model to specialized domains.

\noindent\textit{Challenges of internal knowledge-grounded QA.} A core challenge that comes with QA systems is their tendency to hallucinate information \citep{ZhuangYWS23}, a characteristic that heavily countervails the goal of factual QA.
Thus, there exist many efforts in mitigating hallucinations, making LMs more reliable for QA tasks \citep{ZhouNGD21, JiLFY23}.
We discuss hallucinations thoroughly in \Cref{sec:challenges_factuality}.

\noindent\textbf{External Knowledge:}

\noindent\textit{Description.}
External-knowledge question answering is a text generation task in which a model produces an answer by jointly considering the input question and retrieved or provided external evidence.
External information is provided at inference time, such as a knowledge base, a knowledge graph, or grounded text.
When grounded text is used as the external knowledge source, the task is commonly referred to as reading comprehension.

\noindent\textit{Modeling.}
To leverage external knowledge, most approaches focus on LMs with sufficient context modeling capabilities \citep{TanMLL23}.
They can be prompted to answer a question based on a provided source text \citep{TanMLL23}.
Other knowledge, like knowledge graphs, is leveraged by additional modules, e.g., a graph neural network \citep{LuOZ23}.
\citet{NishidaNSA20} improve the performance of their QA system on specialized domains by stacking a reading comprehension layer and an LM layer on top of BERT \citep{DevlinCLT19a}.
Given specific user requirements, some studies employ custom tools to facilitate or ensure accurate responses.
\citet{ZhuangYWS23} employs 13 external tools, like a code interpreter or a mathematical computer, to enhance the performance of an LLM on the QA task.
Retrieval augmented generation (RAG) combines document vectorization with retrieval \citep{NEURIPS2020_6b493230}.
Queries are embedded, matched against a vector database, and the retrieved context is used to generate grounded responses.
QA datasets with external knowledge are abundant, especially in the subfield of reading comprehension \citep{RogersGA23}.
The widely used SQuAD dataset \citep{RajpurkarZLL16b} contains crowdsourced reading comprehension questions with reference answers supported by a paragraph.
SQuAD 2.0 \citep{RajpurkarJL18} adds 50k unanswerable questions to the base dataset that challenge models to predict whether the generation of an answer is feasible at all.
\citet{DuSC17} and \citet{YuanWGS17} provide a model that can generate synthetic reading comprehension questions.
\citet{GaoBCL19} propose an advanced system with controllable difficulty levels.
It can be used to generate more samples on specific domains if crowdsourcing is not an option.

\noindent\textit{Challenges of external knowledge-grounded QA.} The sub-tasks' difficulty rises with the complexity of the questions.
Despite recent advances, LLMs still exhibit limitations in addressing questions that require complex, multi-step reasoning \citep{TanMLL23}, making effective reasoning with LLMs an active area of research.
Additional challenges arise with non-answerable or non-factoid questions, as LMs are typically trained to always generate some text as an answer \citep{RajpurkarJL18}.
Ideally, a model should be able to tell the user when a competent answer seems unrealistic, given the provided context.
QA datasets like SQuAD \citep{RajpurkarZLL16b} have been criticized because they are overly dependent on the similarity of question/answer sentences rather than on human-like reasoning, meaning they require only superficial reading skills \citep{UtoTS23}.
Datasets like ELI5 \citep{FanJPG19a} can provide alternatives that require more reasoning and multi-step thinking.

\section{Evaluation} \label{sec:eval}
Several metrics have been proposed or adopted from other research fields to evaluate, compare, and discuss text generation systems. 
These metrics serve as proxy measures of desired text-generation characteristics, such as coherence, fluency, and semantic diversity. 
\Cref{tab:table-metrics} shows the summary of all metrics identified.
We identify 16 distinct automated metrics from our systematically retrieved papers (excluding auxiliary papers added during writing) and organize them into three major types: \textit{model-free}, \textit{model-based}, and \textit{human} evaluation.
\textit{Model-free} metrics are usually based on static, rule-based algorithms, such as text overlap (e.g., BLEU \citep{PapineniRWZ02}).
\textit{Model-based} metrics are automated metrics that rely on learned models rather than only surface-level rules.
They include embedding- or encoder-based similarity metrics (e.g., BERTScore \citep{ZhangKWW20a}) or sequence-to-sequence scoring metrics (e.g., BARTScore \citep{YuanNL21c}) to derive a final score.
Another field of evaluation concerns \textit{human evaluation}, which measures characteristics like fluency, faithfulness, coherence, and others in text generation systems \citep{GhoshCLM17, MaSLL18a, QaderJPL18, FengFQQ21, LuWWJ22}.
We also discuss LLM-as-a-judge \citep{ZhengSVF23a}, which uses an instruction-tuned LLM to predict a decision or score.
In the retrieved papers, we find four distinct methods used for human \textit{labeling} (e.g., Likert Scale) and four metrics for annotator \textit{agreement} (e.g., Krippendorff Alpha).

\begin{table*}[t]
    \centering
    \small
            \caption{\label{tab:table-metrics}
        Automated evaluation metrics in the text generation field. We consider systematically retrieved Semantic Scholar documents from 2017 to 2023 after the eligibility criteria (cf. \Cref{sec:manual_assessment}). ``Used'' marks the number of papers that propose, survey, or apply the metric in their publication.
        }
    \renewcommand{\arraystretch}{1.2}
    \begin{tabular}{llp{1.9cm}p{7cm}r}
        \toprule
        \textbf{Type} & \textbf{Category} & \textbf{Metric} & \textbf{Description} & \textbf{Used}\\
        \midrule
        Model-free
        & N-gram & BLEU & Textual overlap between source and reference (precision). & 69\\
        & & ROUGE & Textual overlap between source and reference (recall). & 46\\
        & & METEOR & Textual overlap between source and reference (precision and recall). & 32\\
        & & CIDEr & Measures consensus on multiple reference texts. & 15\\
        & & chrF++ & Character-based F-score computed using n-grams. & 13\\
        & & Dist-n & Measures generation diversity by the percentage of distinct n-grams. & 8\\
        & & NIST & Alters BLEU to also consider n-gram informativeness. & 6\\
        & & Self-BLEU & Measures generation diversity by calculating BLEU between generated samples. & 2\\
        \hdashline
        & Statistical & Perplexity & Fluency metric based on the likelihood of word sequences. & 23\\
        & & Word Error Rate & The rate of words that are different from a reference sequence based on the Levenshtein distance. & 11\\
        \hdashline
        & Graph & SPICE & Measures the semantic similarity of two texts by the distance of their scene graphs. & 6\\
        
        \midrule
        Model-based 
        & Hybrid & BERTScore & Contextual token similarity to measure textual overlap. & 13\\
        & & MoverScore & Uses contextualized embeddings and captures both intersection and deviation from the reference for a similarity score. & 6\\
        & & Word Mover Distance & Distance metric to measure the dissimilarity of two texts. & 2\\
        \hdashline
        & Trained & BLEURT & Models human judgment on text quality. & 4\\
        & & BARTScore & Likelihood metric for assessing candidate consistency with the source or reference text. & 3\\

        \bottomrule
    \end{tabular}
\end{table*}

\subsection{Desired Characteristics in Text Generation}
All metrics are used to estimate the desired characteristics of generated texts.
We identify several characteristics, e.g., fluency and coherence, across the papers we considered.
Although there are many metrics to assess generated texts, they are proxies for \textit{quality}.
Other characteristics, such as bias, faithfulness, and reasoning, also have their own evaluation procedures and will be covered in \Cref{sec:challenges}.

\subsubsection{Quality.}\label{sec:quality} \hfill

\noindent A high-quality text has to fulfill various criteria, such as having a correct grammatical structure or being coherent and fluent.
\citet{Sonntag04} decomposes textual quality into the three general parts: intrinsic, contextual, and representational quality.
We find that this definition partly omits task-specific characteristics.
For example, some tasks like dialogue generation \citep{LiZ23, ZhangWLG20a} specifically require diverse outputs to avoid repetitions and encourage creativity \citep{ChungKA23, AlihosseiniMB19}.
We detail our expanded definitions below.
First, \textit{intrinsic quality} includes the need for a text to be accurate, unbiased, and generated from a data source with a high reputation.
Secondly, \textit{contextual quality} comprises the amount of text being produced, the completeness of the output, the coherence of timeline events, and the relevance of the text to the prompt or document \citep{DengTLX21}.
Third, \textit{representational quality} requires a consistent and concise output that is easily understood.
This also involves the fluency of the output, and the text should be interpretable by the reader.
Lastly, \textit{task-oriented quality} captures requirements specific to the downstream task, such as diversity and creativity in dialogue generation \citep{ChungKA23}, stylistic adherence in creative writing \citep{Peng22}, or faithfulness and coverage in summarization \citep{MaynezNBM20, huang-etal-2024-embrace}.

\noindent\textbf{Challenges of quality evaluation.} Considering these dimensions of text quality (i.e., intrinsic, contextual, representational, and task-oriented), evaluation in text generation research is commonly performed by assessing the similarity of a machine-generated text to a reference text written by humans \citep{PapineniRWZ02, Lin04a}. %
To consider more specific dimensions of quality or additional attributes, but especially task-oriented quality, supplementary metrics are needed.
For example, \citet{JagfeldJV18} measure the textual diversity by the number of unique sentences and words in the output.
\citet{XuLWS17} measure textual novelty by the number of infrequent words in the generated responses, excluding the top 2,000 most frequent words.
Unlike traditional classification and regression tasks, where error rates can be clearly defined, generative tasks have a complex answer space and, in many cases, are subjective. 
Most datasets and tasks (e.g., the summarization dataset XSum \citep{NarayanCL18a} or the paraphrasing dataset ETPC \citep{KovatchevMS18a}) provide only a single reference text.
As a result, automatic evaluation is implicitly bound to one surface realization of the target output.
This limitation causes semantically equivalent but lexically divergent generations to receive low scores due to insufficient token-level overlap, despite being judged as valid by humans.
Consequently, many commonly used metrics, particularly n-gram-based metrics, show limited correlation with human judgments \citep{WahleGR23a, JagfeldJV18}.
Moreover, n-gram-based metrics rely on hard or soft subsequence alignments, which do not measure the interdependence between consecutive sentences well \citep{FabbriKMX21a} and show an unsatisfying correlation with human judgments \citep{GattK18}.
We outline metrics to measure textual quality and other attributes, and to understand the limitations of current evaluation methodologies.

\subsection{Model-Free}
The most used evaluation methods for text generation are model-free metrics because they are easy to adapt and have low computational costs (\Cref{tab:table-metrics}). 
The most common metrics used for evaluating the quality of machine-generated text rely on n-gram-based comparisons, such as BLEU \citep{PapineniRWZ02}, ROUGE \citep{Lin04a}, and METEOR \citep{BanerjeeL05}.
Other metrics compare the semantic differences using graphs (e.g., SPICE \citep{AndersonFJG16}) or the probability of word sequences (e.g., perplexity \citep{JelinekMBB05}).
In general, a metric tries to align well with human judgments, although various human biases and imperfections exist.
As noted earlier, many model-free metrics (e.g., BLEU, ROUGE) do not correlate well with human judgments, as they sometimes assign high scores to low-quality texts \citep{GattK18}.
Still, these metrics are widely used in text generation research as cost-efficient proxies.
Thus, these metrics should be used only under constrained circumstances, i.e., for extractive tasks, and are very limited for tasks with diverse outputs such as abstractive summarization \citep{JADS547}.
We further detail each metric, including its capabilities and limitations.

\label{sec:bleu}
\noindent \textbf{BLEU} is the most employed model-free metric in our review (69 papers) \citep{PapineniRWZ02}. 
It measures the word overlap between a candidate text and one or more references.
The score indicates the precision of the generated text based on the overlapping n-grams.
A high BLEU score indicates that the evaluated text is very similar to the reference in terms of wording.
To mitigate highly precise yet very short texts, a brevity penalty is added.
Notably, BLEU does not account for grammatical errors and only considers the n-grams of a text for comparison.
NIST alters BLEU to also consider the informativeness of n-grams by assigning lower scores to more frequent n-grams than to less frequent ones \citep{Doddington02}.
Self-BLEU measures the diversity of generated samples by applying BLEU to pairs of generations \citep{ZhuLZG18}.
As mentioned before, n-gram-based metrics like BLEU can be susceptible to noisy text \citep{VaibhavSSN19} and are unreliable when used for abstractive tasks \citep{JADS547}.
Thus, their use is limited and best accompanied by additional model-based metrics or human evaluation.

\label{sec:rouge}
\noindent \textbf{ROUGE} measures the recall of n-gram overlap between a generated text and a single reference (46 papers) \citep{Lin04a}.
It is also known as ROUGE-$N$, where $N$ stands for the length of the n-grams used for the calculation (e.g., ROUGE-1 for unigrams, ROUGE-2 for bigrams, etc.).
ROUGE-$L$ specifically considers the longest common subsequence between the generated text and a reference.
Although ROUGE was originally recall-oriented, it is now often reported as ROUGE-F1, which additionally penalizes outputs that achieve high recall by adding irrelevant or excessive text.
As an n-gram-based metric, ROUGE is inherently biased towards lexical similarity \citep{NgA15}.
As a countermeasure, ROUGE-WE \citep{NgA15} adds soft lexical matching through word embeddings (making it a hybrid model-based metric).
Here, the individual vectors of constituent tokens are multiplied to produce the vector for an n-gram.
Just as BLEU, ROUGE is also not reliable for abstractive tasks with diverse outputs \citep{JADS547} and is susceptible to noisy text \citep{VaibhavSSN19}.

\label{sec:meteor}
\noindent \textbf{METEOR} is another overlap metric (32 papers) \citep{BanerjeeL05}.
While BLEU captures precision and ROUGE recall, METEOR combines both into a single metric.
A reordering penalty term punishes texts with the correct words that appear in the wrong order.
Abstractive settings that introduce word-level changes from input to output, such as synonyms, still pose a limitation for the metric.
This is because METEOR is also an overlap-based metric, similar to BLEU and ROUGE.

\label{sec:cider}
\noindent \textbf{CIDEr} is a consensus-based metric that matches a single sentence to a set of reference sentences (15 papers) \citep{VedantamZP15a}.
While originally developed for the image captioning task, CIDEr is often used in various text generation tasks \citep{LiL21a, FabbriKMX21a, LuWWJ22}.
It calculates cosine similarity between 1- to 4-gram co-occurrences and captures precision, recall, grammaticality, and word importance by down-weighting common n-grams.

\label{sec:chrf}
\noindent \textbf{chrF++} is an n-gram-based quality metric that computes n-grams on a character level instead of the token level (13 papers) \citep{Popovic17}.
This makes chrF++ independent from language and tokenization.
It improves correlation with human judgments by incorporating added unigrams and bigrams at the word level.

\label{sec:dist}
\noindent \textbf{Distinct-n} shows the number of distinct n-grams divided by the total number of generated tokens (eight papers) \citep{LiGBG16}.
Researchers use this metric to measure the diversity of a sentence, penalizing sentences with repeated words.
It is not capable of capturing text quality.
Most works use $n\in{1,2,3}$ \citep{NiM18}.

\label{sec:perplexity}
\noindent \textbf{Perplexity} in text generation is defined as the degree of uncertainty of a model in predicting a new sequence (23 papers) \citep{JelinekMBB05}. 
Specifically, perplexity analyzes the likelihood of word sequences, comparing the generated probability distribution of words with the actual input.
A lower perplexity score means the trained model is better at generating with varying input data because it is less perplexed/confused.
Perplexity is only a measure of certainty and, therefore, does not necessarily correlate with the quality of the generated text.
Still, it is a widely used metric that indicates how confident a model is during generation. 

\label{sec:worderrorrate}
\noindent \textbf{Word Error Rate} measures the amount of edits required (substitutions, deletions, insertions) until a sequence equals the reference (11 papers).
Therefore, the Word Error Rate measures the similarity of sentences, indicating textual quality when comparing a result with a reference text.
As the Word Error Rate is a metric commonly used in image captioning tasks, we examined the papers that mention it.
Most mentions of the Word Error Rate in our selection of works either review existing metrics \citep{FatimaIKD22} or apply it to speech recognition \citep{HuangLPH18}, due to its higher sensitivity to specific error types (substitutions, deletions, insertions) compared to more flexible n-gram-based approaches.
It does not capture semantic meaning and therefore cannot distinguish between contextually relevant and irrelevant words.

\label{sec:spice}
\noindent \textbf{SPICE} uses a unique approach to measure semantic similarity by implementing a distance measurement between two texts' scene graphs that encode objects, attributes, and relations (six papers) \citep{AndersonFJG16}.
SPICE ignores fluency, and a highly similar text pair might not be well-formed, but it captures semantic similarity more effectively than n-gram-based metrics. 
This metric is also used for tasks like image captioning because it can capture relations between objects and attributes well \citep{KeneshlooSRR19}.
In the papers we assessed, SPICE is often used alongside other metrics, primarily because it is one of the few model-free metrics that do not rely on n-gram comparisons \citep{LinZSZ20a, LuWWJ22}.

In summary, model-free metrics remain the most commonly used evaluation tools due to their simplicity, low computational cost, and ease of comparison across studies.
However, our analysis confirms that these metrics primarily capture surface-level properties such as lexical overlap, diversity, or prediction confidence, and often fail to reflect semantic adequacy or overall text quality.
Their weak correlation with human judgments, especially for abstractive and open-ended generation tasks, limits their interpretability when used as standalone measures.
Consequently, model-free metrics are best suited to constrained or extractive settings and should be complemented by model-based metrics or human evaluation to yield a more reliable assessment of generated text.

\subsection{Model-Based}

With model-based metrics, we capture the semantic meaning and potentially other attributes of a sentence using neural models such as BERT \citep{DevlinCLT19a}.
Thus, model-based metrics provide a more flexible measurement of textual characteristics.
This relatively novel category of metrics generally outperforms many well-established overlap metrics (e.g., BLEU \citep{PapineniRWZ02, BabakovDLP22a}).
Considering LMs' increasing capabilities, model-based metrics are employed 28 times in the papers from our list, but most studies still include model-free metrics as a widely accepted rule-based measurement, as indicated in \Cref{tab:table-metrics}.
We differentiate between fully end-to-end \textit{trained metrics} that rely on handcrafted features and/or learned embeddings (e.g., BLEURT \citep{SellamDP20}, BARTScore \citep{YuanNL21c}) and \textit{hybrid metrics} that combine trained elements (e.g., contextual embeddings) with computational logic (e.g., BERTScore \citep{ZhangKWW20a}).
Model-based metrics (\textit{trained} or \textit{hybrid}) take advantage of available rating data for their training and should be more robust to distribution drifts than model-free metrics \citep{SellamDP20}. 
\citet{JiLFY23} point out that neural models are prone to producing errors in predicting textual similarity or other attributes.
Thus, relying solely on model-based metrics can propagate errors that degrade the models' accuracy.
Moreover, model-based metrics are inherently prone to insensitivities, biases, and loopholes \citep{HeZWK23a}.
For example, BERTScore \citep{ZhangKWW20a} is confused by truncation errors in summarization.

\label{sec:bertscore}
\noindent \textbf{BERTScore} is the most-used model-based metric in our review (13 papers) \citep{ZhangKWW20a}.
It is a hybrid metric that leverages contextual token similarity to measure the overlap of a text with a reference. 
More specifically, BERTScore uses pairwise cosine similarity to measure the similarity between word embeddings of a source and a reference text.
As it considers precision, recall, and F1 measure, BERTScore can be applied to various text generation tasks and correlates well with human judgments.
The metric is highly expressive and inherently tunable for task-specific properties (e.g., fluency, style, and coherence), though reliability varies across these properties.

\label{sec:moverscore}
\noindent \textbf{MoverScore} uses contextualized representations (e.g., BERT embeddings \citep{DevlinCLT19a}) to measure the similarity between two texts (six papers) \citep{ZhaoPLG19}.
It uses the Word Mover Distance \citep{KusnerSKW15} to determine how much one set of embeddings needs to be changed to transform into the reference embeddings.
Thus, MoverScore considers not only the intersection but also the deviation from the reference.
Sentence Mover's Similarity \citep{ClarkCS19} extends this idea by incorporating sentence embeddings and improving correlation with human judgments.

\label{sec:wordmoverdistance}
\noindent \textbf{Word Mover Distance} calculates the dissimilarity of two texts \citep{KusnerSKW15} using static word representations (e.g., word2vec \citep{MikolovCCD13b}) (two papers).
While Word Mover Distance relies on traditional static word embeddings that can be visualized for interpretability, some researchers prefer MoverScore as a metric due to its greater ability to capture context.

\label{sec:bleurt}
\noindent \textbf{BLEURT} is a fully learned metric that models human judgment on text quality (four papers) \citep{SellamDP20}.
BLEURT is highly adaptable, end-to-end trained, and can be fine-tuned for other tasks.
However, hybrid metrics like BERTScore \citep{ZhangKWW20a} may yield better results with limited training data and do not rely on the assumption that train and test data are similarly distributed.

\label{sec:bartscore}
\noindent \textbf{BARTScore} leverages a text-to-text model to assess text from perspectives like informativeness, coherence, and factuality (three papers) \citep{YuanNL21c}.
While it can also capture precision and recall, BARTScore also considers faithfulness and can be prompted or fine-tuned to fit more specific tasks.
BARTScore++ improves on BARTScore by considering the explicit and implicit error distances when detecting hallucinations \citep{LuDXZ23}.
The application of BARTScore depends on the task, requiring users to decide which variation to choose and how to design the prompt.

In summary, model-based metrics enable a richer evaluation of generated text by capturing semantic and higher-level textual properties that model-free metrics cannot address.
Hybrid approaches such as BERTScore and MoverScore are widely adopted due to their balance between expressiveness and robustness, while fully trained metrics like BLEURT and BARTScore offer stronger alignment with human judgments at the cost of greater sensitivity to biases and distribution shifts.
Despite their overall superior performance, model-based metrics exhibit systematic failure modes and inherited model biases, which limit their reliability when used in isolation.
As reflected in the reviewed studies, they are therefore most effective when combined with model-free metrics to provide complementary and more stable evaluations.

\subsection{Human Evaluation}

\begin{table*}[t]
    \centering
    \small
        \caption{\label{tab:human-metrics}
    Human evaluation methods in the text generation field. We consider systematically retrieved Semantic Scholar documents from 2017 to 2023 after the eligibility criteria (cf. \Cref{sec:manual_assessment}). ``Used'' indicates the number of papers that apply each method.
    }
    \renewcommand{\arraystretch}{1.2}
    \begin{tabular}{llp{1.9cm}p{7cm}r}
        \toprule
        \textbf{Type} & \textbf{Category} & \textbf{Metric} & \textbf{Description} & \textbf{Used}\\
        \midrule
        Human
        & Labeling & Likert Scale
        & Humans can choose on a scale, e.g., from 1 (horrible quality) to 5 (perfect quality).
        & 22\\
        & & Pairwise Comparison
        & Humans choose the best example from two samples.
        & 10\\
        & & Turing Test
        & Can quantify how distinguishable human text is from machine-generated text.
        & 6\\
        & & Binary
        & Humans are answering binary questions with yes or no.
        & 3\\
        & & Best-Worst Scaling
        & From a list of examples, humans are instructed to select the best and worst output.
        & 2\\
        \hdashline
        & Agreement & Krippendorff Alpha
        & Measures the disagreement between annotators for nominal, ordinal, and metric data.
        & 4\\
        & & Fleiss Kappa
        & Measures the agreement on nominal data between a fixed pair of annotators.
        & 4\\
        & & Pearson Correlation
        & Displays the agreement between annotators by measuring linear correlation.
        & 3\\
        & & Spearman Correlation
        & Displays the monotonic relationships on ranked data.
        & 2\\
        \bottomrule
    \end{tabular}
\end{table*}

The standard in assessing the quality, fluency, faithfulness, and coherence of text generation systems is human evaluation. 
Methods for human evaluation can be divided into labeling (e.g., classification) and agreement measurements (e.g., correlations between annotator judgements), which can serve as indicators of the quality of the produced labels.

Usually, human evaluation is performed through crowdsourcing marketplaces (e.g., Amazon Mechanical Turk\footnote{\url{https://www.mturk.com/}}, Prolific\footnote{\url{https://www.prolific.com/}}), where annotators are given the task of labeling samples based on their intuition.
We report the used methods during human annotation in \Cref{tab:human-metrics}.
Ultimately, the choice of evaluation method depends on the task to be solved.
Most used labeling methods are Likert scales (22 papers), e.g., "How coherent is the text from 1 (least coherent) to 9 (most coherent)?".
While the Likert scale is suitable for expressing text quality, ratings are often not explained and don't indicate the problematic text segment \citep{DouFKS22a}.
Other works use pairwise comparisons (ten papers), i.e., directly comparing two systems' outputs to explain comparative improvements \citet{DouFKS22a}.

We find that only a relatively small number of studies report standard agreement metrics, such as Krippendorff's alpha (13 papers).
Especially in complex evaluation setups (e.g., emotion analysis \citep{10534765}), single annotators' judgments can vary substantially.
Without agreement metrics, it is difficult to assess the reliability of the provided labels and to interpret model performance relative to human consistency.
For this reason, we encourage evaluators of text generation systems to report agreement metrics alongside their labels in future work.
While this requires additional annotation effort, agreement metrics indicate the level of human consistency on a task and thus define a meaningful reference point for model performance, particularly when comparing systems on subjective or underspecified evaluations.

Human annotation and evaluation are costly, especially when consulting experts, so they are often avoided or done on a small scale.
\citet{ReiterB09} find that experts and non-experts correlate strongly on features like readability, clarity, and general appropriateness.
\citet{BhandariGAL20a} use the pyramid method \citep{NenkovaP04} to assess automated evaluation metrics themselves through human evaluation.

Some efforts have been made to provide easy-to-use frameworks to increase the effectiveness of human evaluations.
For example, Scarecrow \citep{DouFKS22a} is a framework for crowd-annotating several error types in machine-generated and human-generated text.
Here, ten workers annotate each paragraph to provide an inter-annotator agreement score.
The Multidimensional Quality Metric by \citet{FreitagFGR21} gets judgments for the translation task to improve the quality of human judgments through error analysis.
The metric requires evaluators to identify errors and categorize them into severity levels.
To assess how well a human evaluation and annotation are performed, authors can provide Kendall's Tau or Pearson Correlation.
The TrueSkill algorithm also evaluates human annotations by creating clusters of systems for both quality and naturalness \citep{SakaguchiPV14, GehrmannDER18}.

The LLM-as-a-Judge paradigm uses an instruction-following model as a scalable alternative to human raters by prompting it with the task, example, and the candidate output(s), then extracting either a decision or a score \citep{chatbotarena}.
Methodologically, LLM-as-a-judge is model-based because the evaluator is a learned model. 
Conceptually, however, it approximates human evaluation protocols by prompting a model to produce judgments, preferences, or scores. 
We therefore discuss it in this section as a scalable alternative to human raters, while treating its outputs as model-generated annotations rather than human labels.
Such judges can show high agreement with human preferences but exhibit systematic failure modes such as position/order bias, verbosity/length bias, and occasional self-enhancement or style biases. 
This motivates mitigations like response-order randomization, standardized criteria, and debiasing for confounders such as length \citep{chatbotarena}.
\citet{10534765} find that model-based annotations can be high-quality for simple tasks such as sentiment analysis.
In contrast, human annotations consistently outperform LLM-generated ones in intricate NLP tasks such as emotion classification \citep{10534765}.
In practice, protocols for LLM-as-a-Judge commonly fall into three families: binary judgments (e.g., ``Does the answer satisfy constraint X?''); preference judgments (e.g., ``Which of A vs. B is better under criteria X?''); selection judgments (e.g., ``Which candidate(s) of A, B, C are good answers?'') \citep{li-etal-2025-generation}.

In summary, human evaluation remains the most reliable approach for assessing nuanced and subjective aspects of text generation.
However, its high cost and inter-annotator variability limit scalability and reproducibility.
The limited reporting of agreement metrics further complicates the interpretation of results.
While LLM-as-a-Judge methods offer a scalable alternative and show promising alignment with human preferences in simple settings, they exhibit systematic biases and reduced reliability for complex tasks.
Consequently, human evaluation should be complemented with agreement analysis and, where appropriate, automated judging to ensure both validity and scalability.

\section{Related Challenges} \label{sec:challenges}
Although text generation has seen much progress concerning its proposed techniques, related challenges remain that have not yet been solved.
Aside from the specific challenges and limitations in \Cref{sec:textgen}, we identify nine main challenges related to text generation.
The challenges are \textit{bias}, \textit{reasoning}, \textit{factuality}, \textit{misuse}, \textit{datasets}, \textit{interpretability}, \textit{transparency}, \textit{privacy}, and \textit{computing}.
This section describes each challenge while surveying their state-of-the-art mitigation methods and remaining research gaps.
We note that some of the terms appear anthropomorphic (e.g., hallucination, reasoning).
As these terms are well-established and widely used in the scientific literature, we also use them.
However, we aim to explain each term by a neutral description.

\subsection{Bias}

Bias, in the most general sense, describes systematically erroneous output that portrays scenarios distortedly \citep{ShengCNP19}.
In text generation, this distortion often leans towards or against certain demographic groups or concepts and amplifies biases in the training sets \citep{SunGTH19}.
As bias can occur in different shapes, this problem touches several other areas, such as toxicity in text generation \citep{ZhengSVF23a} and sentiment-controlled text generation \citep{GhoshCLM17, HuangZJS20, Kawamae23, LiuSLS21}.
Bias towards groups or concepts is a pressing problem for machine-generated text \citep{ShengCNP19, DhamalaSKK21}, and its mitigation can reduce the subliminal influence on perceived stereotypes by users of such systems.

While the characteristics of bias are manifold, we identify three main types: \textit{exposure bias}, \textit{allocation bias}, and \textit{representation bias}.
First, \textit{exposure bias} occurs when a model is only exposed to the training data distribution instead of its prediction \citep{KeneshlooSRR19}.
To avoid \textit{exposure bias}, \citet{KeneshlooSRR19} remove the ground-truth dependency during training and use only the model distribution to minimize the loss function. 
Reinforcement learning concerning a metric like ROUGE \citep{Lin04a} can mitigate \textit{exposure bias} \citep{WangYTZ18, JinG22}.
A Generative Adversarial Network can be employed as the generator constantly relies on its output during training \citep{ShenLC22}.
Second, \textit{allocation bias} appears when models perform better on data associated with groups or situations overrepresented in the training data \citep{SunGTH19}.
\citet{ZhaoWYO18} show how gender-specific \textit{allocation bias} can be mitigated by creating a second dataset with a swapped gender though it doubles the dataset size and is expensive to create.
Third, \textit{representation bias} appears when associations between groups with certain concepts are captured in word embeddings or model parameters.
Thus, the representation (e.g., the embedding) reflects the bias \citep{SunGTH19}.
An intuitive way to tackle \textit{representation bias} is to debias the embeddings directly by regularization \citep{HuangZJS20}.
This comes with a trade-off between generation quality and control because it can impose the target attribute on the LM on improper positions \citep{GuFMW22, HuangZJS20}.
Debiasing can be effective while only fine-tuning 1\% of parameters on an unbiased dataset, making learned techniques relatively cheap to adopt \citep{GiraZL22a}.
Notably, \citet{SchickUS21} show that pre-trained LMs are capable of recognizing their own biases to a considerable degree.
Thus, they can be prompted to self-diagnose their output.

The quantification of bias is especially relevant when considering an LM for real-world applications and determining the effectiveness of mitigation techniques.
BOLD \citep{DhamalaSKK21} and FairPrim \citep{FleisigAAB23} both are helpful datasets to measure various biases (profession, gender, race, religious belief, and political ideology).
BOLD metrics \citep{DhamalaSKK21} further assess the texts' sentiment, toxicity, regard, psycholinguistic norms, and gender polarity.
Notably, not all biases are easy to measure with current metrics due to either a lack of labeled data specific to the bias type or corresponding noise in the supposedly bias-free data.

\subsection{Reasoning} \label{sec:challenges_reasoning}
Reasoning describes the capability of a text generation system to computationally infer its choices and writing from the source context logically and sensibly.
A model should be able to make an understandable decision and explain its reasoning to a human.
Several works point out that, even though recent years yielded significant improvements, LMs have difficulties with such commonsense inference \citep{LinZSZ20a, ZellersHBF19a}.
Thus, eliciting reasoning with text generation is an active research topic.
LLMs struggle with sequential decision-making that requires common-sense planning, as shown by evaluating their performance on reinforcement learning benchmarks \citep{10.1145/3643795.3648387}.
The lack of reasoning capabilities comes from misunderstanding facts in the source context and is also referred to as intrinsic hallucinations \citep{JiLFY23} (more details in \Cref{sec:challenges_factuality}).
Text generation systems require reasoning capabilities for complex tasks like story generation \citep{LiangTLZ23} or question answering \citep{FanJPG19a}.

One popular approach to improve reasoning is Chain-of-Thought Prompting (CoT) \citep{WeiWSB24}.
Here, the model is prompted to think step by step and output its thought process in natural language.
It improves performance for arithmetic, commonsense, and symbolic reasoning tasks.
Multi-agent systems like Exchange-of-Thought \citep{YinSCG23} maintain the reasoning capabilities of CoT \citep{WeiWSB24} while allowing for an iterative self-interaction that improves the output text.
The approach imitates a multi-turn conversation between experts on a task.
Nevertheless, the effectiveness of multi-agent systems is debated, as strong prompting of a single model can lead to similar task performance \citep{wang-etal-2024-rethinking-bounds} and prolonged agentic interaction increases the risk of performance collapse \citep{becker2025stayfocusedproblemdrift}.
\citet{ChenJX20} survey another method to improve reasoning capabilities via path-finding strategies on a knowledge graph.
We also find considerable research related to testing a model's reasoning capabilities.
\citet{StolfoJSS23} explain how to specifically test the robustness of mathematical reasoning by grounding a behavioral analysis in a causal graph.
The CommonGen dataset \citep{LinZSZ20a} requires relational reasoning and compositional generalization. 
The HellaSwag dataset \citep{ZellersHBF19a} can also be used to test a model's ability of generative commonsense reasoning.

\subsection{Hallucinations} \label{sec:challenges_factuality}
Hallucinations describe outputs that are factually wrong or unfaithful to the source.
We distinguish between \textit{intrinsic} and \textit{extrinsic hallucinations}, which refers to what content the generated text contradicts to \citep{JiLFY23}.
\textit{Intrinsic hallucinations} contradict the source content, so they are unfaithful to the training data or input text.
\textit{Extrinsic hallucinations} describe outputs that can neither be supported nor contradicted by the source content \citep{MaynezNBM20}.

The main reasons for hallucinations stem from the inherent sampling in statistical language models and from the source data, because diversity-valuing tasks like dialogue generation inherently generate text that deviates from the factual information in the source data \citep{JiLFY23}.
Bias in the training data and text generation models can also lead to factually wrong outputs \citep{QaderJPL18, JiLFY23}.
Improving factuality in text generation makes systems more reliable in real-world applications (e.g., chat support assistants and summarization systems).

Overall, models pre-trained on large amounts of data tend to hallucinate less \citep{MaynezNBM20, ZhouNGD21}.
\citet{PrabhumoyeHZB21} mitigate hallucinations in document-grounded text generation by implementing a selective attention variant for the documents' contextualized representations.
Faithfulness to a source can be improved by representing the knowledge in a simpler relational format and verifying it against relation tuples from the source or reference \citep{GoodrichRLS19}.

Other works employ mechanisms to detect hallucinations in existing text.
Measuring hallucinations with LMs can be challenging they are designed to predict texts of high linguistic quality (e.g., fluency, grammaticality) rather than factual content \citep{GattK18}.
In addition, most automatic metrics do not correlate with faithfulness or factuality, making them often unreliable.
Human evaluation on factuality datasets \citep{AbhishekSSS22, ZhouNGD21} is the preferred method \citep{MaynezNBM20, GoyalD20a, FalkeRUD19, JiLFY23}.
\citet{GoyalD20a} present a textual entailment system that can fact-check generated text and localize the factual error.
The model-based metric BARTScore++ can detect hallucinations to a certain degree because it imitates human-like error analysis \citep{LuDXZ23}.
\citet{FabbriWLX22} propose QAFactEval to evaluate factual consistency for the QA task.
\citet{MadaanPPS21} propose a model to automatically generate counterfactual text that can be leveraged to create a synthetic test set on factuality.

\subsection{Misuse}

It is important to name the adversarial risks of text-generative models. 
Only when knowing how these systems can be misused can countermeasures be researched.
This is also referred to as threat modeling \citep{ZellersHRB20}, i.e., identifying threats and vulnerabilities from an adversary's point of view.
We identify three main categories of how text generation systems are misused currently or in the near future.
These are \textit{non-disclosed}, \textit{misaligned}, and \textit{adversarial} usage.

Machine-generated systems can pose a threat if their usage is \textit{not disclosed} properly.
Convincing machine-generated text poses a significant challenge to academic integrity \citep{CottonCS24}. %
Additionally, highly controllable text generation systems can flood social media \citep{ZellersHRB20, TourilleSP22} or product pages \citep{AdelaniMFN20} with intentionally biased and non-factual information (fake news, propaganda, fake product reviews).
The main mitigation methods for undisclosed usage concentrate on discriminating machine-generated text from human-written text \citep{TourilleSP22, WahleRKG22b, WahleRMG21b, FagniFGM21, MunirBSS21}, although attribution alone does not verify a text's factuality \citep{SchusterSSB20a}.
Classifiers for machine-generated text suffer from regular false positive detections for many reasons, ranging from shortness to incoherence \citep{JawaharAL20}.
\citet{PerkinsRPM24} rely on well-trained academic staff using intricate annotation software to identify generated content.
A RoBERTa-based detector achieves notable results on a dataset comprising human- and machine-generated data \citep{maktabdar2025detection}. 
Still, reliable detection is especially difficult when advanced prompting techniques are used \citep{PerkinsRPM24}.
An adversary using multiple text generators within the same document might pose an additional challenge for detectors \citep{UchenduMLZ21}.
\citet{RodriguezHGS22} suggest that paragraph-level detectors can be used to detect the tampering of full-length documents.
GLTR utilizes the fact that human writing includes more sequentially improbable tokens in a sequence than machine text and highlights tokens probabilities \citep{GehrmannSR19a}.
\citet{ZhongTXW20} specifically assess the factual structure of a text to detect machine samples.
\citet{10594194} employ a hybrid detector, merging conventional TF-IDF strategies with Bayesian classifiers, stochastic gradient descent, categorical gradient boosting, and 12 instances of LMs.
\citet{ni2025mvanmultiviewattentionnetworks} develop a multi-view attention mechanism to detect fake news that attends to both the semantics and propagation structure of social media content.

Users of text generation systems might have harmful intentions and specifically tailor prompts that lead to text generation that is \textit{not aligned} with human ethics or safety regulations.
To mitigate \textit{misaligned} use, many models like GPT-4 \citep{AchiamAAA24}, LLaMA 2 \citep{TouvronLIM23}, or Gemini \citep{GeminiTeamABW23} include automated censorship mechanisms that block potentially harmful content.
For example, Gemini \citep{GeminiTeamABW23} leverages advanced Chain-of-Thought recipes to align with safety policies.
GPT-4 \citep{AchiamAAA24} provides an additional reward signal during RLHF fine-tuning \citep{OuyangWJA22} that targets correct behavior, such as refusing to give harmful responses.
However, these mechanisms can be tricked.
For example, \citet{JiangXNX24} leverage the poor performance of many modern LMs in recognizing ASCII art to jailbreak safety-aligned LMs.

A more technical misuse of machine-generated text is \textit{adversarial}. 
As text classification systems are sensitive to certain words, an adversary could intentionally generate text including certain words, ultimately hindering the effectiveness of classification models \citep{MoscaARG22, GaoLSQ18, JinJZS20}.
Due to the high instruction-following capabilities of modern LMs \citep{OuyangWJA22}, an additional risk arises from prompt injections.
Prompt injections describe secretly hidden text snippets (prompts) in any content used by an LM-based application \citep{GreshakeAME23}.
They can lead to the extraction of sensitive information from LMs (e.g., personal data, credentials) \citep{GreshakeAME23}.
Prompt injections can also be used for phishing, scams, or masquerading, and some systems might be vulnerable to unwanted API calls.
To avoid tailored texts tricking classifiers, \citet{MoscaARG22} propose a logits-based metric that captures words with a suspiciously high impact on a classifier.
The risks of text-generative models are abundant, and mitigation methods must be tailored to each problem.
Meanwhile, we emphasize raising awareness of the diverse problems to foster the development of mitigation methods.

\subsection{Datasets}
The data scale has increased significantly since contemporary LMs need to be pre-trained on large amounts of web-scraped data.
As corpora grow, finding and filtering open-source and high-quality data becomes more challenging, which might encourage exploring licensed data for training (e.g., transcribed YouTube videos).
Also, many datasets and their sources are unstandardized \citep{FatimaIKD22}.

\citet{FleisigAAB23} make several suggestions on how to improve the quality of datasets.
Datasets should clarify the potential risks (e.g., toxicity \citep{GehmanGSC20a}) of the provided machine-generated text.
With high-quality annotations, authors can measure or mitigate a diverse set of fairness-related harms, and identify the demographic groups of annotators \citep{FleisigAAB23}.
In addition, datasets should account for annotator disagreement, providing individual annotators' judgments rather than a final score.
Context-dependent harms can only be identified when the context is provided in the dataset (e.g., perceived author of text, application), and a diverse set of annotators can ensure that various demographic groups are represented.

A key problem related to data scarcity concerns multilingual applications.
Text generation works best for languages encountered frequently during pre-training.
However, most text generation research takes place in the United States, followed by economically strong countries such as China, the United Kingdom, and India \citep{RamosMR23, AbdallaWRN23}.
Thus, most research is performed on corresponding corpora, mostly in English or Chinese.
Low-resource languages remain underexplored in machine-generated text, and multi-lingual models must be specifically trained to fit the niche \citep{ShahamHAS24}.

To enhance a model's performance on scarce data like low-resource languages, researchers apply variations to the training process.
\citet{ShahamHAS24} show that just 40 multilingual samples in an English tuning set improve multi-lingual instruction following.
This also generalizes to unseen languages during fine-tuning, indicating that including two to four languages in the training set significantly improves cross-lingual generalization.
Datasets should not be used repeatedly as a benchmark, as overfitting reduces their efficiency long term \citep{FleisigAAB23}.
Thus, there exists a constant need for new datasets. 
One example dataset is XWikiRef \citep{TaunkSPS23}, a Wikipedia summarization dataset on eight low-resource languages across five domains (i.e., books, films, politicians, sportsmen, and writers).
We highlight the importance of data-efficient methodologies to improve the utility of text generation systems across languages, dialects, and sociolects.

\subsection{Interpretability}
A text generation model is interpretable when the user can understand why a certain output was produced.
Modern LMs inherently do not provide intuitive insight into how textual outputs are produced, as their weights and representations appear mathematically complicated \citep{LuMXQ22}.

Interpretability is a desired goal when solving tasks because, without the reasoning behind the final output being clear, the correctness of the output can not be easily assessed, and users need to rely solely on trusting the model. 
Moreover, researchers rely on the intricate analysis of existing models to propose novel ideas.

\citet{SinghIGC24} point out that, overall, two aspects of interpretation research need to be explained.
First, we require an intricate explanation of an LM's behavior.
This comprises local characteristics like feature attribution and data grounding and global characteristics like attention head importance and data influence.
Second, a clear explanation of used datasets is required.
This includes interactive clarification using natural language and aiding data analysis.
\citet{TenneyWBB20a} propose a tool to visualize an LMs' behavior on the architectural level.
They visualize the sentence embeddings, classification probabilities, attention, and confusion matrix.
By providing the information in a visual format that is easy to understand, one can directly assess how an LM behaves when specific data points are altered.
Ultimately, tools like this provide a meaningful step towards interpretability, as models' inner processes usually remain hidden and their disclosure requires time-consuming development of custom solutions.

\subsection{Transparency}
Transparency involves two major aspects.
First, transparency refers to the lucid explanation and documentation of proposed models and datasets.
This includes the research along with their underlying methodologies, which many newly released models like GPT-4 \citep{AchiamAAA24} and Gemini \citep{GeminiTeamABW23} only provide superficially.
As the quality of AI-authored scientific work remains subpar relative to human-authored research \citep{amirjalili2024exploring}, especially in terms of factual accuracy and complex aspects of authorship, the disclosure of its use in academic writing also becomes essential.
The indiscriminate use of machine-generated text during training introduces irreparable defects into the resulting models, given the recursive incorporation of machine-generated data \citep{shumailov2024ai}.
So-called model collapse could be avoided by fully disclosing the use of AI, although this can be considered an ambitious solution.

Since big tech companies increasingly lead and, consequently, shift NLP research directions, the discussion of potential risks (e.g., toxicity, factuality) has become essential in recent years, especially when releasing LMs \citep{AbdallaWRN23}.
To research and address the emerging challenges, the transparent reporting of model characteristics is of key importance.

For this, \citet{MitchellWZB19} provide an extensive and standardized model card that can be shipped together with novel LMs.
However, corporate organizations often have a monetary interest in releasing new models, and fully disclosing all model details can conflict with a company's interest in not having their product replicated or released to an open-source audience.
\citet{AczelW23} describe the transparent and credible usage of text-generative models in scientific writing as threefold.
First, researchers should disclose the model and its training data used.
In addition, the generated text, or a summary of it, should be available to the readers.
Second, the authors should specify for which aspects of a scientific work (outline, writing, code) a model was used.
Third, researchers are required to always verify generated text which includes a check on plagiarism.
Fulfilling all aspects described by \citet{AczelW23} can be difficult, but the guidelines give an idea of the ideal process for documenting AI usage.
\citet{WahleRMM23} propose AI Usage Cards, a shorter card that can be included in research papers to document how text generation models are used for the work.

\subsection{Privacy}
The data collection process raises privacy concerns as there is apprehension about companies using human inputs to their system to continuously train their models \citep{WuDN24}.
Modern LMs require large amounts of data crawled from the web for their pre-training \citep{TouvronLIM23} and it is hard to control whether personal data is being used. 
These models have been found to output memorized passages from their training data \citep{YuanCRI22, CarliniTWJ21}. 
Thus, LMs could also leak private information they encountered during training \citep{LiTZW21, GreshakeAME23}.

\citet{LiGFX23} show how multi-step jailbreaking prompts can be formulated to expose private information like names or emails, even bypassing safety modules employed in applications like ChatGPT \citep{YuanCRI22}. 
Examples like this show that privacy-related research about text generation is crucial.

\citet{YueILK23} use differential privacy to mathematically guarantee that private information can neither be generated nor reconstructed from the output.
The key idea of differential privacy is that including or excluding a single individual's data should not significantly affect the model's output.
By understanding the sensitivity regarding single data points, the algorithm knows how much noise to add to a stochastic gradient descent algorithm.
Once the noise has been added, further data processing will not diminish privacy protection.
This property is crucial because it ensures privacy is not compromised through additional analysis.
However, their mathematical guarantee of privacy protection comes with a trade-off regarding output quality.
Notably, tight differential privacy harms small classes like minority groups during training as rarely mentioned groups could be nullified in the process.
Thus, one needs to find the right balance between privacy and quality in real-world applications.
\citet{SongS19} develop an auditing model that lets users check whether their data has been used to train a text generation model.

\subsection{Computing}
LMs have been growing increasingly large \citep{ZhaoZLT23a} and with more data and a higher number of model parameters, training times also surge. 
This requires more computing and energy to publish a novel state-of-the-art model.
Nowadays, this becomes a problem for small companies and research organizations as only big tech companies like OpenAI, Google, or Meta have the resources to train a competitive language model from the ground up \citep{AbdallaWRN23}.
Larger models also have a higher environmental cost, emitting more carbon during training than smaller models \citep{TouvronLIM23, FabbriWLX22}.
Reducing the computational requirements of recent LMs could potentially lower their environmental impact due to lower power consumption.

The scaling laws from \citep{KaplanMHB20a} suggest that larger models will continue to perform better than smaller ones while reaching the same level of performance with fewer optimization steps.
They also find that a model's performance depends strongly on the number of parameters, dataset size, and the amount of computing, while architectural parameters like layer depth or width are not as important.
Thus, making text generation models lightweight and improving computing times is an active research topic.
\citet{LiuOPS23} propose a transformer with ternary (i.e., -1, 0, 1) or binary (i.e., 0, 1) weights, which facilitate multiplication-free computations.
Division-of-Thoughts distributes a reasoning process across a locally deployed, smaller-scale LM and cloud-based LLMs \citep{10.1145/3696410.3714765}, reducing reasoning time and API costs.
EdgeShard deploys LLMs across distributed services, accounting for device heterogeneity and bandwidth limitations \citep{10818760}.
Their variation of BART \citep{LewisLGG20a} is up to 16 times more efficient than its real-valued counterpart while achieving close to normal performance.
However, binarization and ternarization require bit-packing to have actual memory savings and need dedicated hardware support for real-time acceleration.
\citet{LiL21a} propose prefix-tuning, a light-weight alternative to fine-tuning where only 0.1\% of a models parameters are optimized.
Their model performs comparable to a full-data setting and is even better in low-data settings and extrapolation on unseen topics.
Approaches to reduce LMs' environmental impact are often not included in the works we read.
We identify that NLP-focused research on reducing environmental harm is scarce but necessary if the field wants sustainable future research.

\section{Epilogue} \label{sec:epiloge}

In this literature review, we covered 257 publications, focusing on recent developments in text generation from the Transformer architecture in 2017 through December 2025.
In that context, we systematically identified and discussed the core tasks and sub-tasks in the field, assessed how text generation systems are evaluated, and outlined relevant challenges that researchers can address in the near and mid-term future.
To summarize this work, we revisit our initial research questions.\\[5pt]
\noindent \textbf{What constitutes the task of text generation? What are the main sub-tasks?}\\[3pt]
\noindent We categorized text generation into five main sub-tasks: \textit{open-ended text generation}, \textit{summarization}, \textit{translation}, \textit{paraphrasing}, and \textit{question answering}.
We identified prominent sub-tasks and unique challenges to display recent research within each of these.
\textit{Open-ended text generation} comprises open-domain applications such as prompted instruction-following systems, story generation, and dialogue generation.
Open-ended tasks are inherently challenged by trying to consistently model long contexts, especially when considering sub-tasks that are lexically and semantically diverse, like dialogue generation or story generation.
In \textit{summarization}, we focused on single documents, multiple documents, and dialogues.
Recent studies concentrate on sub-tasks like multi-document summarization and dialogue summarization, which exhibit higher contextual complexity.
While sentence-level \textit{translation} is a well-established task, document-level translation poses a chance to mitigate limitations like word ambiguities within or between documents.
\textit{Paraphrasing} research shifts from uncontrolled generation towards more controllable scenarios.
Here, we noticed a lack of aligned paraphrasing datasets that provide labels for such controllable attributes.
\textit{Question answering} relies on internal and external knowledge sources and is challenged by reasoning, non-answerability, and non-factoid questions.\\[5pt]
\noindent \textbf{How are text generation systems evaluated? What are the concomitant limitations?}\\[3pt]
\noindent We identified two main evaluation methodologies: model-free and model-based.
While model-free evaluations are based on static and rule-based algorithms, model-based ones leverage word or sentence embeddings of pre-trained models to produce a score.
Most model-free metrics rely on n-gram overlap, but their correlation to quality as humans perceive it varies.
Moreover, some textual characteristics, like factuality or bias, are difficult to measure using model-free metrics, which is why we emphasized the need for additional model-based metrics and human evaluation.
In addition to quality, model-based metrics can capture attributes like factuality or writing style by fine-tuning or prompting.
However, as neural models are subject to errors, model-based metrics might introduce various biases and fail to reliably quantify these attributes, while also lacking interpretability.
LLM-as-a-Judge provides a scalable evaluation alternative with strong agreement to human judgments on simple tasks, but suffers from systematic biases and underperforms humans on complex NLP tasks.
While many studies use human labels for evaluation, they often omit measures of inter-annotator agreement or label reliability.
We encourage more researchers to consider agreement and reliability in the design phase of human studies.
For this, we point out helpful metrics, methodologies, and labeling tools to improve the quality of produced annotations.\\[5pt]
\noindent \textbf{What are the open challenges in text generation?}\\[3pt]
\noindent We covered \textit{bias}, \textit{reasoning}, \textit{hallucinations}, \textit{misuse}, \textit{datasets}, \textit{interpretability}, \textit{transparency}, \textit{privacy}, and \textit{computing}.
In this work, we identified three prominent \textit{bias} types relevant to text generation: exposure-, allocation-, and representation bias.
Recent works are concerned with bias measurement and mitigation in generated texts.
External modules that quantify and detect \textit{hallucinations} can provide transparency and improve factuality in text generation.
Specific prompting techniques like Chain-of-Thought stimulate language models' \textit{reasoning} capabilities, but they require an intricate prompt design.
We outlined various scenarios of how text generation could potentially be \textit{misused}.
These scenarios are categorized into non-disclosed, misaligned, and adversarial usage.
Additionally, existing \textit{datasets} for text generation tend to disregard low-resource languages and are sometimes not standardized into a uniform format or lack metadata like inter-annotator agreement scores.
\textit{Interpretability} in text generation should consider both the used models and associated datasets.
Differential \textit{privacy} can mitigate the risk of leaking private information from training data at the cost of output quality.
To make capable language models more accessible to the community, we find incipient research about how to make the \textit{computing} requirements of text generation systems lightweight without sacrificing substantial amounts of performance.
This involves both the training process and the inference of
models.\\[5pt]
\noindent \textbf{What are prominent research directions in text generation?}\\[3pt]
Recent publications have focused more on improving the intrinsic (e.g., bias, factuality) and contextual (e.g., relevance, completeness) quality of the machine-generated text.
We suspect research in these areas will be particularly fruitful, and we highlighted various directions accordingly.
For example, we identified that many bias mitigation methods depend on careful data selection and augmentation while only working for specific bias types.
Research towards more universal mitigation techniques, possibly leveraging self-aware language models to quantify their own biases, would provide an impactful contribution to the community.
Improvements in factuality are challenged by ponderous evaluation.
To foster additional research in that area, we suggested investigating the effective quantification of hallucinations using model-based methods.
Aside from prompting strategies, enhancing language models' reasoning capabilities could be considered during a model's training instead of during inference.
With language model applications permeating people's everyday lives, subordinate problems like its misuse become apparent.
To ensure the safety and responsible use of these text-generative models, safety-aligned systems should be able to detect when they are about to be exploited by misaligned or adversarial prompts.
In addition, responsible individuals and companies are required to establish safeguards to prevent the abuse of text-generation systems.
Excluding private information from both the training process of language models and the generation of text poses another opportunity for research.
Compared to more established tasks like uncontrolled paraphrasing or single-document summarization, contemporary proposals suggest that the community's general interest shifted towards more specific scenarios (e.g., controlled paraphrasing) and sub-tasks that are contextually more complex (e.g., dialogue summarization).
We suggest that such scenarios will be more prominently researched in subsequent works.
Additionally, we identified the potential to use large pre-trained models that can operate across a variety of domains and tasks.
Future work could focus on more effectively leveraging the internal knowledge of such models for tasks like question answering while avoiding hallucinated content.
In summary, we discussed several sub-tasks of text generation that leave space for future research and outlined key challenges that can be targeted in the near and mid-term future.

\subsection{Limitations}\label{sec:limitations}
Text generation can involve various other modalities that are not immediately natural language.
For example, methods include table generation \citep{LuWWJ22}, table-to-text generation \citep{WangWAY20}, sql-to-text generation \citep{XuWWF18}, or text generation from an abstract meaning representation \citep{MagerANS20}.
In this work, we focus on text generation from natural language input because modalities such as images \citep{TianZXW24} require specialized techniques and are therefore out of scope for this review.
Some recently published works may not have passed the filters of our systematic review because they might not have had enough time to accumulate enough citations.
Ultimately, this is a limitation of our bottom-up approach, as some subfields might not be captured by this systematic search.
We tried to mitigate this by mentioning related subfields where appropriate.
We release the manual relevance judgments to the public through our GitHub repository\textsuperscript{\ref{github_project}}.
When retrieving papers, we rely on Semantic Scholar's set of indexed papers and ranking algorithm, which might lead to some papers being left out, although Semantic Scholar's coverage, particularly in AI and NLP, is high.
We detail the relevance function used by Semantic Scholar for proper documentation of the process.

\begin{acks}
This work was supported by the Lower Saxony Ministry of Science and Culture and the VW Foundation.
This work was funded by the Deutsche Forschungsgemeinschaft (DFG, German Research Foundation) – 564661959.
\end{acks}

\bibliography{mainbib2,additional2, resubmission}
\bibliographystyle{apalike}

\clearpage

\clearpage
\onecolumn
\hypertarget{annotation}{}
\pagestyle{empty}
\lstset{
  basicstyle=\footnotesize\ttfamily,
  breaklines=true,
  breakatwhitespace=false,
  columns=flexible,
  numbers=none
}

\definecolor{Primary}{RGB}{59, 130, 246}    %
\definecolor{PrimaryDark}{RGB}{30, 64, 175} %
\definecolor{LightBg}{RGB}{239, 246, 255}   %
\definecolor{TextDark}{RGB}{31, 41, 55}     %
\definecolor{TextMuted}{RGB}{107, 114, 128} %

\begin{tikzpicture}[remember picture, overlay]
  \fill[Primary] ([xshift=0cm,yshift=0cm]current page.north west) rectangle ([xshift=\paperwidth,yshift=-0.4cm]current page.north west);
\end{tikzpicture}

\vspace{0.8cm}
\begin{center}
  {\fontsize{22}{26}\selectfont\sffamily\bfseries \textcolor{PrimaryDark}{CiteAssist}}\\[0.2em]
  {\Large\sffamily\scshape \textcolor{TextMuted}{Citation Sheet}}\\[0.8em]
  {\small\sffamily Generated with \href{https://citeassist.uni-goettingen.de/}{\textcolor{Primary}{\texttt{citeassist.uni-goettingen.de}}}
  \CiteAssistCite{}
  }\end{center}

\begin{center}
\vspace{1em}
\begin{tikzpicture}
\draw[Primary, line width=0.6pt] (0,0) -- (\textwidth,0);
\end{tikzpicture}
\vspace{1.2em}
\end{center}

\begin{tcolorbox}[enhanced,
                 frame hidden,
                 boxrule=0pt,
                 borderline west={2pt}{0pt}{Primary},
                 colback=LightBg,
                 sharp corners,
                 breakable,
                 fonttitle=\sffamily\bfseries\large,
                 coltitle=Primary,
                 title=BibTeX Entry,
                 attach title to upper={\vspace{0.2em}\par},
                 left=12pt]
\lstset{
    inputencoding = utf8,  %
    extendedchars = true,  %
    literate      =        %
      {á}{{\'a}}1  {é}{{\'e}}1  {í}{{\'i}}1 {ó}{{\'o}}1  {ú}{{\'u}}1
      {Á}{{\'A}}1  {É}{{\'E}}1  {Í}{{\'I}}1 {Ó}{{\'O}}1  {Ú}{{\'U}}1
      {à}{{\`a}}1  {è}{{\`e}}1  {ì}{{\`i}}1 {ò}{{\`o}}1  {ù}{{\`u}}1
      {À}{{\`A}}1  {È}{{\`E}}1  {Ì}{{\`I}}1 {Ò}{{\`O}}1  {Ù}{{\`U}}1
      {ä}{{\"a}}1  {ë}{{\"e}}1  {ï}{{\"i}}1 {ö}{{\"o}}1  {ü}{{\"u}}1
      {Ä}{{\"A}}1  {Ë}{{\"E}}1  {Ï}{{\"I}}1 {Ö}{{\"O}}1  {Ü}{{\"U}}1
      {â}{{\^a}}1  {ê}{{\^e}}1  {î}{{\^i}}1 {ô}{{\^o}}1  {û}{{\^u}}1
      {Â}{{\^A}}1  {Ê}{{\^E}}1  {Î}{{\^I}}1 {Ô}{{\^O}}1  {Û}{{\^U}}1
      {œ}{{\oe}}1  {Œ}{{\OE}}1  {æ}{{\ae}}1 {Æ}{{\AE}}1  {ß}{{\ss}}1
      {ẞ}{{\SS}}1  {ç}{{\c{c}}}1 {Ç}{{\c{C}}}1 {ø}{{\o}}1  {Ø}{{\O}}1
      {å}{{\aa}}1  {Å}{{\AA}}1  {ã}{{\~a}}1  {õ}{{\~o}}1 {Ã}{{\~A}}1
      {Õ}{{\~O}}1  {ñ}{{\~n}}1  {Ñ}{{\~N}}1  {¿}{{?\`}}1  {¡}{{!\`}}1
      {„}{\quotedblbase}1 {“}{\textquotedblleft}1 {–}{$-$}1
      {°}{{\textdegree}}1 {º}{{\textordmasculine}}1 {ª}{{\textordfeminine}}1
      {£}{{\pounds}}1  {©}{{\copyright}}1  {®}{{\textregistered}}1
      {«}{{\guillemotleft}}1  {»}{{\guillemotright}}1  {Ð}{{\DH}}1  {ð}{{\dh}}1
      {Ý}{{\'Y}}1    {ý}{{\'y}}1    {Þ}{{\TH}}1    {þ}{{\th}}1    {Ă}{{\u{A}}}1
      {ă}{{\u{a}}}1  {Ą}{{\k{A}}}1  {ą}{{\k{a}}}1  {Ć}{{\'C}}1    {ć}{{\'c}}1
      {Č}{{\v{C}}}1  {č}{{\v{c}}}1  {Ď}{{\v{D}}}1  {ď}{{\v{d}}}1  {Đ}{{\DJ}}1
      {đ}{{\dj}}1    {Ė}{{\.{E}}}1  {ė}{{\.{e}}}1  {Ę}{{\k{E}}}1  {ę}{{\k{e}}}1
      {Ě}{{\v{E}}}1  {ě}{{\v{e}}}1  {Ğ}{{\u{G}}}1  {ğ}{{\u{g}}}1  {Ĩ}{{\~I}}1
      {ĩ}{{\~\i}}1   {Į}{{\k{I}}}1  {į}{{\k{i}}}1  {İ}{{\.{I}}}1  {ı}{{\i}}1
      {Ĺ}{{\'L}}1    {ĺ}{{\'l}}1    {Ľ}{{\v{L}}}1  {ľ}{{\v{l}}}1  {Ł}{{\L{}}}1
      {ł}{{\l{}}}1   {Ń}{{\'N}}1    {ń}{{\'n}}1    {Ň}{{\v{N}}}1  {ň}{{\v{n}}}1
      {Ő}{{\H{O}}}1  {ő}{{\H{o}}}1  {Ŕ}{{\'{R}}}1  {ŕ}{{\'{r}}}1  {Ř}{{\v{R}}}1
      {ř}{{\v{r}}}1  {Ś}{{\'S}}1    {ś}{{\'s}}1    {Ş}{{\c{S}}}1  {ş}{{\c{s}}}1
      {Š}{{\v{S}}}1  {š}{{\v{s}}}1  {Ť}{{\v{T}}}1  {ť}{{\v{t}}}1  {Ũ}{{\~U}}1
      {ũ}{{\~u}}1    {Ū}{{\={U}}}1  {ū}{{\={u}}}1  {Ů}{{\r{U}}}1  {ů}{{\r{u}}}1
      {Ű}{{\H{U}}}1  {ű}{{\H{u}}}1  {Ų}{{\k{U}}}1  {ų}{{\k{u}}}1  {Ź}{{\'Z}}1
      {ź}{{\'z}}1    {Ż}{{\.Z}}1    {ż}{{\.z}}1    {Ž}{{\v{Z}}}1  {ž}{{\v{z}}}1
  }
\begin{lstlisting}
@article{becker2026text,
  title     = {Text Generation: A Systematic Literature Review of Tasks, Evaluation, and Challenges},
  author    = {Becker, Jonas and Wahle, Jan Philip and Gipp, Bela and Ruas, Terry},
  journal   = {Journal of Artificial Intelligence Research},
  volume    = {86},
  articleno = {48},
  numpages  = {49},
  month     = aug,
  year      = {2026},
  doi       = {10.1613/jair.1.22258},
  url       = {https://jair.org/index.php/jair/article/view/22258}
}
\end{lstlisting}
\end{tcolorbox}

\vspace{0.8em}
\begin{tcolorbox}[enhanced,
                 frame hidden,
                 boxrule=0pt,
                 borderline west={2pt}{0pt}{Primary},
                 colback=LightBg,
                 sharp corners,
                 breakable,
                 fonttitle=\sffamily\bfseries\large,
                 coltitle=Primary,
                 title=Online Access,
                 attach title to upper={\vspace{0.2em}\par},
                 left=12pt]

\renewcommand{\arraystretch}{1.5}
\begin{tabular}{@{}p{0.25\textwidth}@{}p{0.75\textwidth}@{}}

\textbf{\sffamily JAIR} & 
\begin{minipage}[t]{0.72\textwidth}
\href{https://jair.org/index.php/jair/article/view/22258}{\color{Primary}https://jair.org/index.php/jair/article/view/22258}
\end{minipage}\\
\end{tabular}

\end{tcolorbox}

\vfill
\begin{tikzpicture}
\draw[Primary!40, line width=0.4pt] (0,0) -- (\textwidth,0);
\end{tikzpicture}
\begin{center}
\small\sffamily\textcolor{TextMuted}{Generated \today}
\end{center}

\end{document}